\documentclass[journal]{IEEEtran}
\usepackage[raggedright]{titlesec}

\usepackage{hyperref}
\usepackage{multirow}

\usepackage{cite}
\ifCLASSINFOpdf
   \usepackage[pdftex]{graphicx}
\else
   \usepackage[dvips]{graphicx}
\fi
\usepackage{amsmath}
\usepackage{amsfonts}
\usepackage{stfloats}

\usepackage[ruled,vlined]{algorithm2e}
\usepackage{algorithmic}
\usepackage{array}
\usepackage{caption}
\usepackage{subfigure}

\begin{document}
%
% paper title
% Titles are generally capitalized except for words such as a, an, and, as,
% at, but, by, for, in, nor, of, on, or, the, to and up, which are usually
% not capitalized unless they are the first or last word of the title.
% Linebreaks \\ can be used within to get better formatting as desired.
% Do not put math or special symbols in the title.
\title{Corner Detection Based on Multi-directional Gabor Filters with Multi-scales}
%
%
% author names and IEEE memberships
% note positions of commas and nonbreaking spaces ( ~ ) LaTeX will not break
% a structure at a ~ so this keeps an author's name from being broken across
% two lines.
% use \thanks{} to gain access to the first footnote area
% a separate \thanks must be used for each paragraph as LaTeX2e's \thanks
% was not built to handle multiple paragraphs
%
\IEEEoverridecommandlockouts

\author{\IEEEauthorblockN{Huaqing~Wang\IEEEauthorrefmark{1},
		Junfeng~Jing\IEEEauthorrefmark{2*}, Ning~Li\IEEEauthorrefmark{3},  Weichuan~Zhang\IEEEauthorrefmark{4} and
		Chao~Liu\IEEEauthorrefmark{5}}\\
	\IEEEauthorblockA{
		\IEEEauthorrefmark{1}School of Electronic Information, Xi'an Polytechnic University, No.19 Jinhua South Road, Xi'an, 710048, China,\\
		\IEEEauthorrefmark{2}Xi'an Polytechnic University Branch of Shaanxi Artificial Intelligence Joint Laboratory, Xi'an, 710048, China,\\
		\IEEEauthorrefmark{3}School of Electronic Information, Xi'an Polytechnic University, No.19 Jinhua South Road, Xi'an, 710048, China,\\
		\IEEEauthorrefmark{4}CSIRO Data61, PO Box 76, Epping, NSW 1710, Australia,\\ \IEEEauthorrefmark{5}School of Electronic Information, Xi'an Polytechnic University, No.19 Jinhua South Road, Xi'an, 710048, China.}\thanks{\IEEEauthorrefmark{2*} Junfeng~Jing's email is Jingjunfeng0718@sina.com.}}

\maketitle

% As a general rule, do not put math, special symbols or citations
% in the abstract or keywords.
\begin{abstract}
Gabor wavelet is an essential tool for image analysis and computer vision tasks. Local structure tensors with multiple scales are widely used in local feature extraction. Our research indicates that the current corner detection method based on Gabor wavelets can not effectively apply to complex scenes. In this work, the capability of the Gabor function to discriminate the intensity changes of step edges, L-shaped corners, Y-shaped or T-shaped corners, X-shaped corners, and star-shaped corners are investigated. The properties of Gabor wavelets to suppress affine image transformation are investigated and obtained. Many properties for edges and corners were discovered, which prompted us to propose a new corner extraction method. To fully use the structural information from the tuned Gabor filters, a novel multi-directional structure tensor is constructed for corner detection, and a multi-scale corner measurement function is proposed to remove false candidate corners. Furthermore, we compare the proposed method with twelve current state-of-the-art methods, which exhibit optimal performance and practical application to 3D reconstruction with good application potential.
\end{abstract}

% Note that keywords are not normally used for peerreview papers.

% For peer review papers, you can put extra information on the cover
% page as needed:
% \ifCLASSOPTIONpeerreview
% \begin{center} \bfseries EDICS Category: 3-BBND \end{center}
% \fi
%
% For peerreview papers, this IEEEtran command inserts a page break and
% creates the second title. It will be ignored for other modes.
\IEEEpeerreviewmaketitle

\section{Introduction}
% The very first letter is a 2 line initial drop letter followed
% by the rest of the first word in caps.
%
% form to use if the first word consists of a single letter:
% \IEEEPARstart{A}{demo} file is ....
%
% form to use if you need the single drop letter followed by
% normal text (unknown if ever used by the IEEE):
% \IEEEPARstart{A}{}demo file is ....
%
% Some journals put the first two words in caps:
% \IEEEPARstart{T}{his demo} file is ....
%
% Here we have the typical use of a "T" for an initial drop letter
% and "HIS" in caps to complete the first word.
The kernel of the Gabor wavelet resembles the two-dimensional receptive field profile of a simple cell in the mammalian cortex, with ideal spatial localization and direction selectivity characteristics, which gives it the best localization in the spatial and frequency domains. Gabor features have been successfully applied to biometrics due to their robustness to local distortions caused by differences in illumination, expression, and pose. The biometric field mainly includes face recognition\cite{daisyinvestigation01,rathika2020recognition03,huang2022deep}, iris recognition\cite{daugman2009iris013}, palmprint verification\cite{sardar2022secure,shao2021deep,luo2016local026}, and human gait recognition\cite{arshad2022multilevel,lopez2021survey}. Furthermore, Gabor filters are also commonly used in general image structure information catching, such as corner detection\cite{zhang2014corner,quddus1998corner31,gao2004corner32,gao2007multiscale33}, texture analysis\cite{grigorescu2002comparison,chen2022improved}, and image matching\cite{nunes2017local,tahir2022improving}.

A robust keypoint~\cite{9866553,10026417,JING2023118673,JING2022259,8883063,9234393,WANG2020107299,ZHANG2017193} is a key to outstanding local features of an image. For example, image matching, object tracking\cite{loncomilla2016object}, and motion estimation all require the critical pre-processing operation of detecting keypoints. Moreover, detecting keypoints is a hot spot in academic research. Moravec\cite{moravec1980obstacle} analyzes the corners in the image with distinct intensity variations in each direction, which are highly identifiable, and therefore launches a study on corner detection. Inspired by Moravec's experimental results, Harris et al.\cite{harris1988combined} proposed the Harris detector to detect corners, which can derive first-order functions in the horizontal and vertical directions to describe corner features in images. Because of its simplicity and high efficiency, the Harris detector has become one of the most successful methods and has been widely used. However, the detection system based on a single scale leads to the loss of certain salient points and acceptance of false points\cite{lee1995multiscale}. The distinction between edges and corners cannot be effectively represented by two-directional derivatives\cite{noble1988finding}.

Therefore, many detectors have been presented to achieve a more robust and accurate detection of key points to address these difficulties. A series of detectors\cite{wang2022efficient,mikolajczyk2004scale,lowe2004distinctive,zeng2022small,alcantarilla2012kaze,miao2013interest,duval2015edges15} exploited the concept of multi-scale detection to obtain a wide range of scale features as much as possible. Meanwhile, contour-based methods\cite{zhang2014corner,zhang2015contour,zhang2015laplacian,zhang2019discrete24}, and template-based methods\cite{smith1997susan,rosten2008faster,shui2013corner,xia2014accurate29} have also been presented. The results of a previous phase of image edge detection are strongly reliant on the outcomes of contour-based approaches for keypoint recognition since they analyze the shape changes on the edge contours derived from an edge detector's input image. The template-based methods used predefined parameterized templates to fit a small patch of an image, yet the incomplete coverage of some critical positions affects its actual detection performance. Recently, Zhang and Sun\cite{zhang2020corner30} utilized anisotropic Gaussian directional derivatives (SOGGDD) to depict intensity changes of several general models (such as L-shaped corners, star-shaped corners, T-shaped corners, Y-shaped corners, X-shaped corners, and step edges). In addition, they also discovered that an anisotropic Gaussian kernel is a helpful tool for determining the difference between edges and corners, further illuminating the fact that a corner's intensity variation in most directions is generally high but not always in all directions. So multi-direction detection is as necessary as multi-scale detection.

Gabor transform can be used to analyze the local properties of signals as well as for multi-scale analysis. Gabor wavelets can be precisely localized in a planar domain consisting of time and frequency. Besides, they directly reveal the shape and orientation of the regional structure. Due to advanced performance, applying the Gabor wavelet in traditional corner detection is competitive. Quddus and Fahmy\cite{quddus1998corner31} used the difference between two low-pass Gabor filters with different bandwidths to extract the corners. The filtering is done iteratively until the change in the output is below a certain threshold. Gao et al.\cite{gao2004corner32} deconstructed the original image using Gabor wavelets at various scales and directions. They chose the value orthogonal to the gradient direction as the location of the detection corner. 

In \cite{gao2007multiscale33}, the magnitudes of the log-Gabor wavelet transform are formulated into two directional structure tensors for corner detection, and the smaller eigenvalue of this matrix was taken as a judgment basis. Zhang et al.\cite{zhang2014corner} proposed the use of the imaginary part of the Gabor filter to process the image's pixels on the edge contours. The normalized magnitude responses in each direction are summed, distinguishing the corners from the smoothed edge contours. However, the detectors, as mentioned earlier, contained unavoidable blemishes. Gao et.al \cite{gao2007multiscale33} indicated that the modified Gabor corner detector performed well only on simple synthetic images and was not robust enough for the detection of natural images. The effectiveness of corner detection by contour-based corner detectors\cite{zhang2014corner} depends on the ability to detect contours. 

The Gabor transform can select many texture features. Still, the Gabor kernel is highly non-orthogonal and can cause redundancy in the coefficients\cite{po2006directional}, which makes it difficult for Gabor detectors to distinguish corners from other pixels in an image. In \cite{ren2020contour,field1987relations}, they embedded Gabor and Log-Gabor\cite{gao2007multiscale} wavelets in the corner detection algorithm for capturing the local grayscale variation and geometric structure of the image to achieve accurate detection of corners. Kumar et al.\cite{kumar2002defect} extracted gray change information using the imaginary part of Gabor filter (IPGF) to extract feature points by making full use of the change magnitude in different directions. The IPGF detector improves the reliability of detection by utilizing the feature information of contours in the image and the gray change information. Zhao et al.\cite{zhao2019interest} used a multi-scale Gabor filter based on this, smoothed the image, and then obtained the normalized information entropy at different scales as the feature of corner points.

Moreover, we found that a series of Gabor detectors in the corner extraction process weighted summation of local structural information when analyzing the image leads to a partial loss of structural information, which in turn affects the localization and extraction accuracy of detected corners. So far, there is no corner detector that combines the excellent characteristics of Gabor wavelets with efficient, robust, and accurate corner detection. This motivates us to carry out further research.

In this work, we focus on some shortcomings of past Gabor corner detectors and improve them. A corner detection method based on Gabor features with an expandable multi-directional structure tensor is proposed, a measure that can utilize the multi-scale and multi-directional structure information in images. In addition, we propose a multi-scale corner measurement function to reduce noise effects and suppress unexpected mutations, which can remove spurious candidate corners. The proposed corner detection method is compared with 13 representative corner detectors. It is shown that the proposed method improves the detection precision, corner localization precision, affine transformation, illumination variation, viewpoint variation, and noise relative to the current state-of-the-art corner detection methods. Furthermore, we apply our algorithm to 3D reconstruction and compare it with the reconstruction results of current fashionable algorithms, and we find that our detector has the best performance.

The other parts of the paper are arranged as described below. In Section 2, we illustrate the basic concepts of the Gabor wavelets. Meanwhile, the various corner models are distinguished by using anisotropic Gabor filters. Section 3 presents an extended multi-directional tensor of structure based on Gabor features and a new algorithm for multi-scale corner measurement and detection. In Section 4, experiments are given to evaluate each detectors, and experiments in 3D reconstruction and image matching have demonstrated the excellent performance of our detector. Finally, the work of this paper is summarized in Section 5.

\section{Related Work}
In this section, we review the basic concepts of the Gabor wavelets. Meanwhile, the various corner models will be distinguished using anisotropic Gabor filters.

\subsection*{2.1 Gabor Wavelets}
In the spatial domain, a two-dimensional Gabor filter is a Gaussian kernel function modulated by a sine plane wave, consisting of orthogonal real and imaginary parts, which can be used separately or in combination. Gabor filters are directional and frequency selective and have the best joint resolution in both spatial and frequency domains. Therefore, Gabor filters are used as bandpass filters to eliminate noise and retain accurate structural information. We apply multi-scale analysis\cite{daubechies1990wavelet0G5} to Gabor filters at different frequencies, rendering them versions of each other at different scales. The general formula diagram of the two-dimensional Gabor filter is as follows\cite{kamarainen2006invariance0G6}
\begin{align}\label{eq_gabor}
\centering
\psi (x,y)&=\frac{f^{2}}{\pi \gamma \eta }e^{-(\frac{f^{2}}{\gamma ^{2}}{x}'^{2}+\frac{f^{2}}{\eta ^{2}}{y}'^{2})}e^{j2\pi f{x}'}\\
{x}'&=xcos\theta +ysin\theta \nonumber\\
{y}'&=-xsin\theta +ycos\theta \nonumber
\end{align}
where $f$ is the filter's central frequency, $\theta$ is the rotation angle of the Gaussian major axis and the plane wave, $\gamma$ is the sharpness along the major axis, and $\eta$ is the sharpness along the minor axis (perpendicular to the wave). Note that the filter's center is defined in polar coordinates with parameters $(f,\theta ) $. In the given form, the aspect ratio of the Gaussian is $\lambda =\eta /\gamma$.

\begin{figure}[htbp]
	\centering
	\includegraphics[width=3.2in]{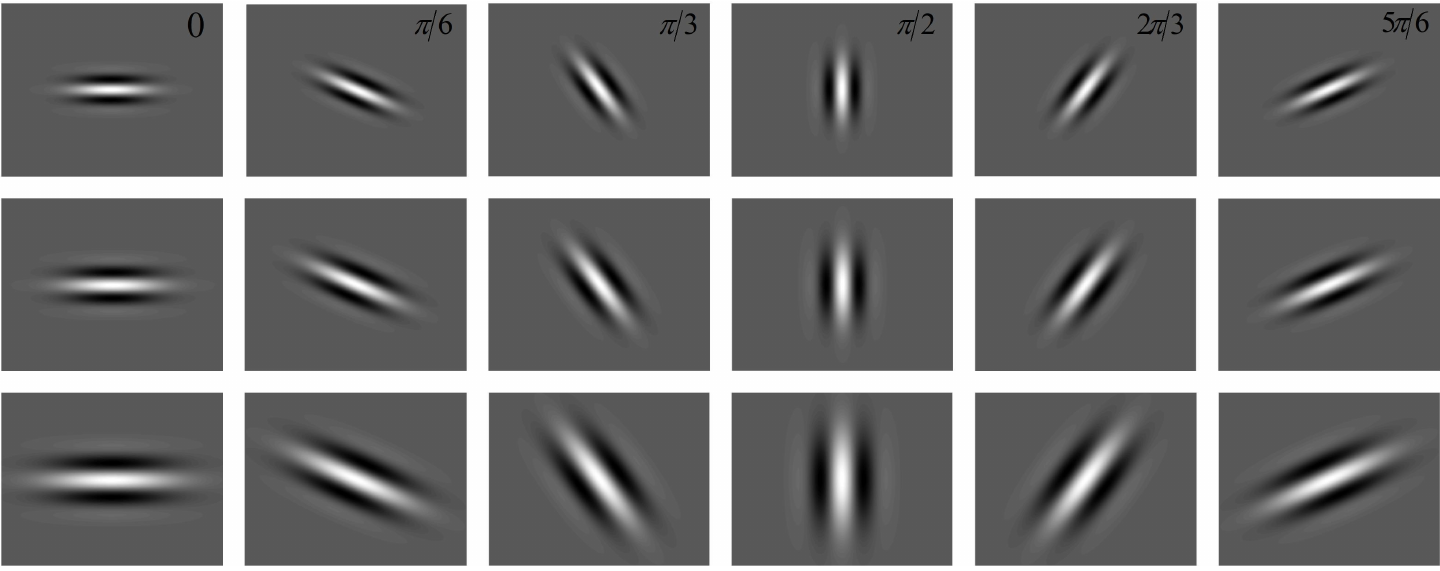}
	\caption{The imaginary parts of Gabor functions for three different scales and six different directions.}
	\label{fig_gaborbank}
\end{figure}

\begin{figure}
	\centering
	\includegraphics[width=3.2in]{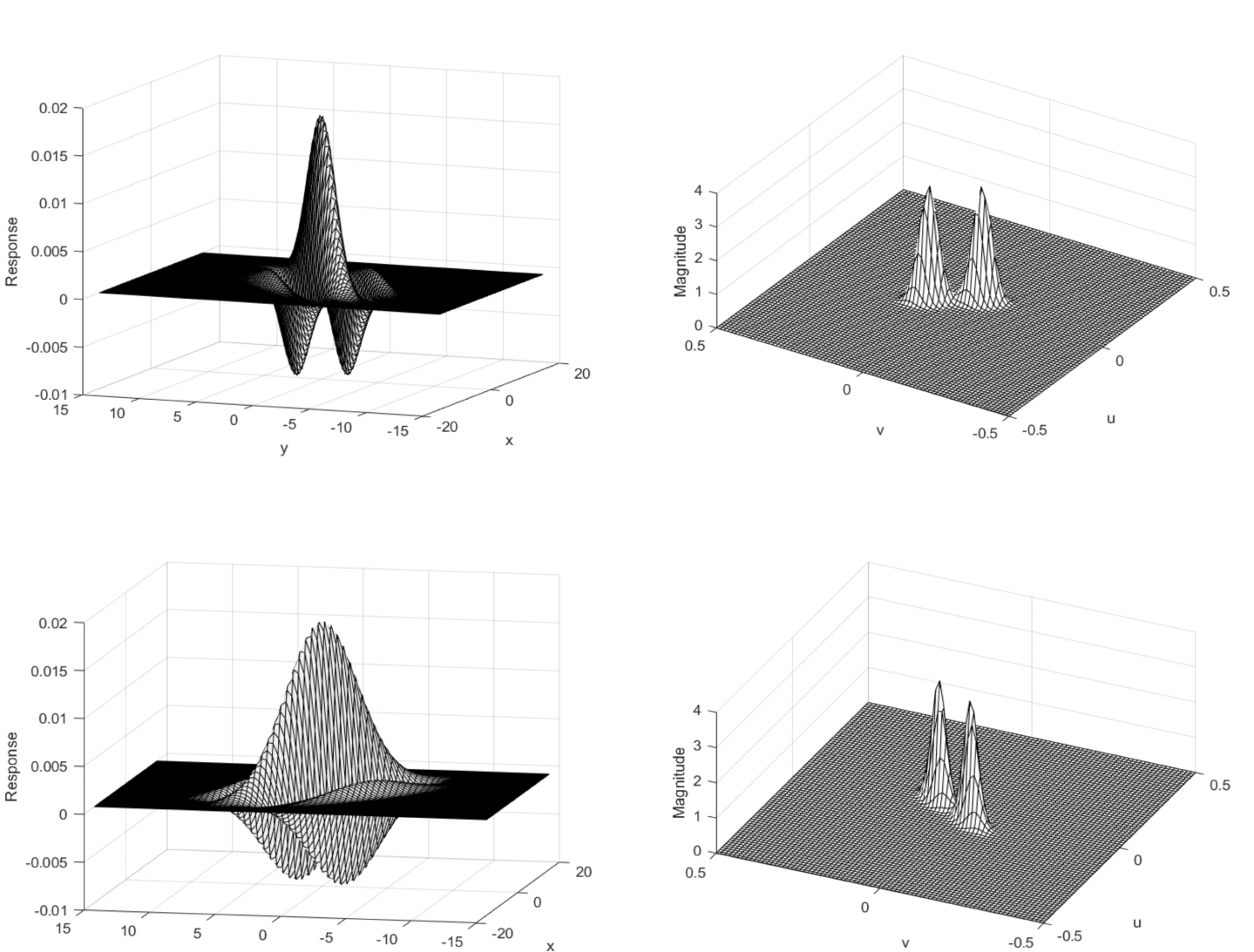}
	\caption{Examples of the imaginary parts of 2-D Gabor filters in spatial and frequency domains. (a) $\gamma =0.6, \eta =1.2, f= 0.2, \theta=\frac{{5\pi }}{6} $. (b) $\gamma =0.6, \eta =1.2, f= 0.2, \theta=\frac{{\pi }}{6} $.}
	\label{fig_frequency}
\end{figure}

The Gabor kernels $\psi _{f,\theta }(x,y)$ are self-similar, and all Gabor kernels are generated by the rotation and scaling of master wavelets (parameters $(f,\theta ) $), so features can be extracted at different scales and in different directions in the frequency domain. Each two-dimensional Gabor filter is obtained by multiplying the Gaussian core function and the sine plane wave, in which $e^{j2\pi f{x}'}$ represents the oscillating part of the Gabor core. Fig.\ref{fig_gaborbank} shows the results of imaginary spectral dimensional transformations in six directions at three scales of the Gabor kernels, with parameters as follows: $\gamma =0.6, \eta =1.2, f\in \left \{ 0.25,0.2,0.15 \right \}$. As shown in Fig.\ref{fig_frequency}, we select the Gabor kernel function to represent the optimal spatial frequency, spatial location, and directional selectivity in local structural information. The Gabor kernel function is convoluted with the test image I (x, y) to obtain the filtered image $Gabor{I_{f,\theta }}$. The formula is as follows
\begin{align}\label{eq_GaborI0}
Gabor{I_{f,\theta }} = I\left( {x,y} \right) * {\psi _{f,\theta }}\left( {x,y} \right)
\end{align}

\subsection*{2.2 The Ability to Distinguish Step Edges and Different Shapes of Corners}
It has been demonstrated that anisotropic Gaussian filters may describe the intensity change disparities between step edges and corners\cite{zhang2020corner30}. Inspired by this, we employ the imaginary part of the Gabor function to differentiate between different corner models. As we all know, the imaginary part of the Gabor function is excellent in extracting image-critical structure information. This Gabor kernel can be expressed in the spatial domain as
\begin{align}\label{eq3}
{\phi _{f,{\theta _k} }}\left( {x,y} \right) &=\frac{f^{2}}{\pi \gamma \eta }e^{-(\frac{f^{2}}{\gamma ^{2}}{x}'^{2}+\frac{f^{2}}{\eta ^{2}}{y}'^{2})}sin(2\pi f{x}')\\
{x}'&=xcos{\theta _k} +ysin{\theta _k} \nonumber\\
{y}'&=-xsin{\theta _k} +ycos{\theta _k} \nonumber\\
{\theta _k} &= {{k\pi } \mathord{\left/{\vphantom {{k\pi } {K,k = 0,1, \ldots ,K - 1}}} \right.\kern-\nulldelimiterspace} {K,\ k = 0,1, \ldots ,K - 1}} \nonumber
\end{align}

For the input test image $I\left( {x,y} \right)$, the mathematical expression of the corresponding image after filtering by the Gabor filter is
\begin{align}\label{eq_GaborI}
Gabor{I_{f,{\theta _k} }} = I\left( {x,y} \right) * {\phi _{f,{\theta _k} }}\left( {x,y} \right)
\end{align}

\begin{figure}
	\centering
	\includegraphics[width=2.0in,height=2.0in]{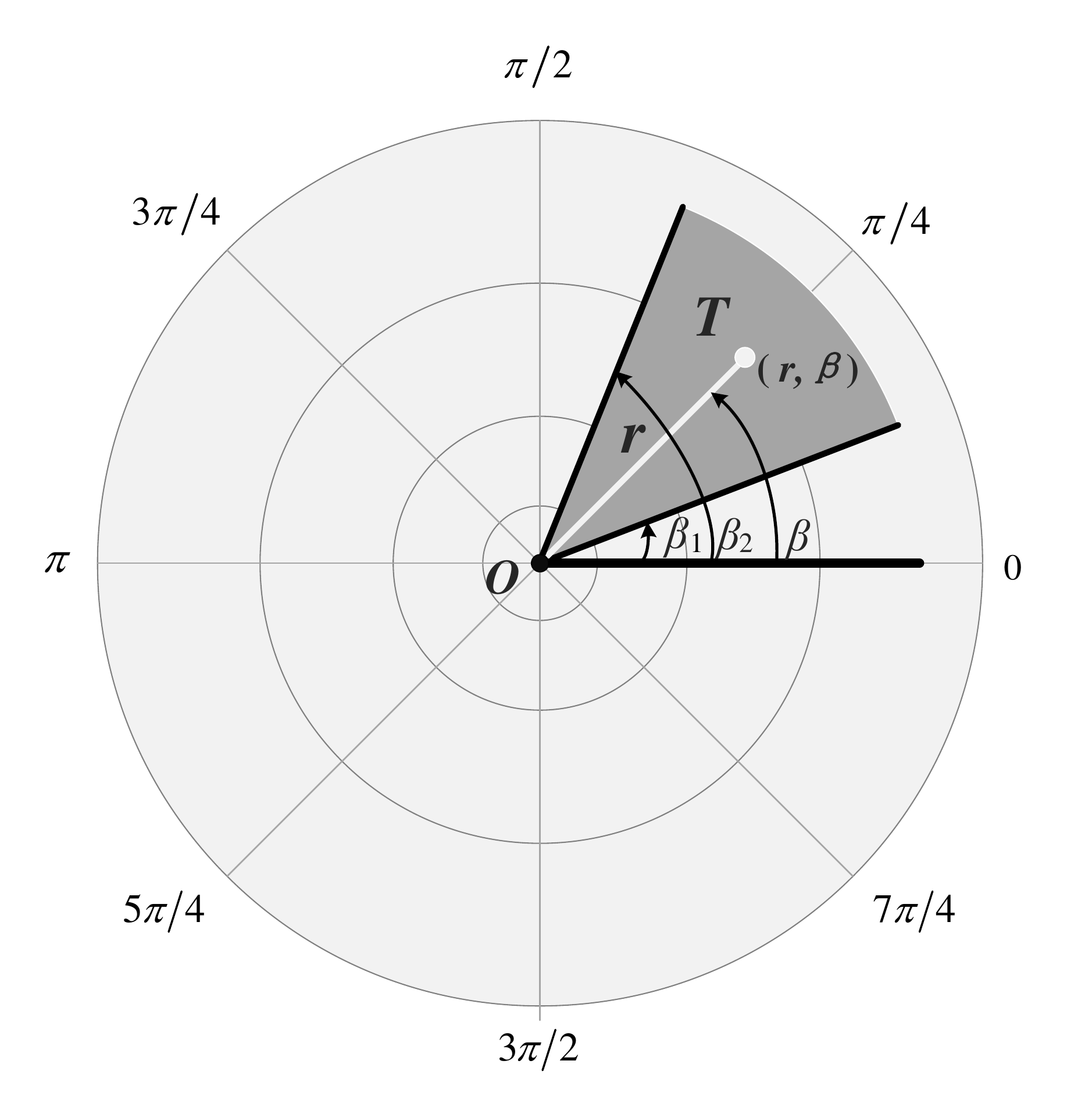}
	\caption{Examples of a basic corner model in the polar coordinate system.}
	\label{fig_models}
\end{figure}

In the polar coordinate system, a point function in a wedge-shaped region can be represented as
\begin{align}
{\zeta _{{\beta _1},{\beta _2}}}\left( {r,\beta } \right) = \left\{ {\begin{array}{*{20}{c}}
	T ,&\text{if } 0\leq r< \infty ,\beta _{1}\leq \beta \leq \beta _{2},\beta _{2}-\beta _{1}\neq \pi\\
	0 ,&\text{otherwise}
	\end{array}} \right.
\end{align}
where $T$ means the gray value of the corner model, $\beta$ represents the corner model's angle in polar coordinates, $r$ represents the corner model's polar diameter, and ${\beta _2}$ and ${\beta _1}$ represent the upper and lower limits of this corner model in the polar coordinate system (with a difference of no more than 180 degrees), respectively. The corner model in polar coordinates is shown in Fig.\ref{fig_models}. The vertices of this wedge-shaped corner model are set as the vertices of the corners and used as a base corner model for corner detection in images. The corner and vertex detection model was proposed in \cite{deriche1993computational}.

In this paper, a series of base corner models are summarized, which are represented by
\begin{align}\label{eq_models}
{h_{\left( {{T_i},{\beta _i}} \right)}}\left( {r,\beta } \right) = \sum\limits_{i = 1}^s {{T_i}} {\zeta _{{\beta _i},{\beta _{i + 1}}}}\left( {r,\beta } \right),
\end{align}
where ${{\zeta _{{\beta _1},{\beta _2}}}{(r,\beta )}}$ is the generalized representation of the point function in the corner model region, $s$ means the number of corner model regions, and ${T_i}$ means the grayscale value of the $i$-th corner model region. When Eq.\ref{eq_models} means an L-shaped corner, the corresponding conditions are $s=2$ and ${\beta _2} - {\beta _1} \ne \pi$. When Eq.\ref{eq_models} means a step edge, the corresponding conditions are $s=2$ and ${\beta _2} - {\beta _1} = \pi$. When Eq.\ref{eq_models} means a Y- or T-shaped corner, the corresponding condition is $s=3$. When Eq.\ref{eq_models} means a star-shaped corner, the corresponding condition is $s=5$. When Eq.\ref{eq_models} means an X-shaped corner, the corresponding state is $s=4$. In this work, the Gabor filter ${{\phi _{f,{\theta _k} }}\left( {x,y} \right)}$ in Eq.\ref{eq3} is used to smooth L-shape, Y-shape, T-shape, X-shape, and star-shape corner models in the polar coordinate system. The Gabor kernel filter for each corner model is given as follows
\begin{align}
\xi _{f}\left ( {\theta _k}  \right )=\iint_{\mathbb{R}^{2}}^{}\zeta _{\beta _{1},\beta _{2}}\left ( r,\beta  \right )\phi _{f,{\theta _k} }\left ( -r,-\beta  \right )rdrd\beta ,
\end{align}
where $\mathbb{R}^{2}$ is the 2D real space and $\phi _{f,{\theta _k} }\left ( r,\beta  \right )$ is the Gabor filter in the polar coordinate system.

\begin{figure*}
	\centering
	\includegraphics[width=7.2in,height=2.2in]{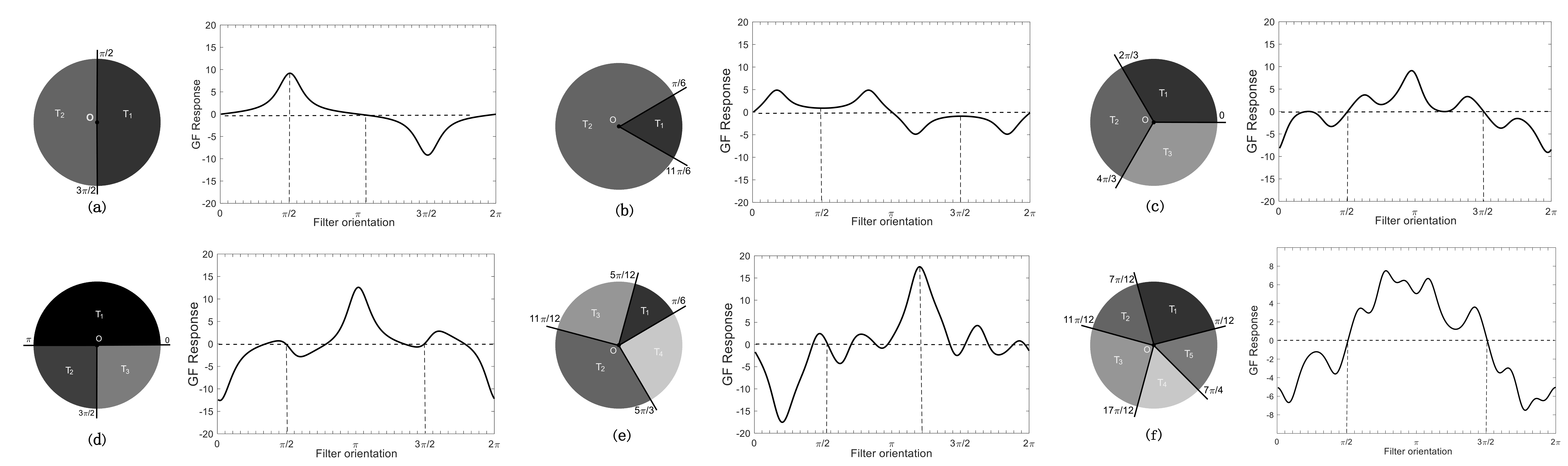}
	\caption{The step edge, L-type corner, Y-type corner, T-type corner, X-type corner, and star-type corner models are shown in (a)-(f) ( gray value $T_{1} = 50, T_{2} = 100, T_{3} = 150, T_{4} = 200, and \ T_{5} = 120$ ). Their corresponding Gabor filtered responses($ f = 0.2,\gamma  = 0.6,\eta  = 1.2 $) are shown beside the appropriate corner model respectively.}
	\label{fig_corner-models}
\end{figure*}

\begin{figure*}
	\centering
	\includegraphics[width=7.2in,height=1.15in]{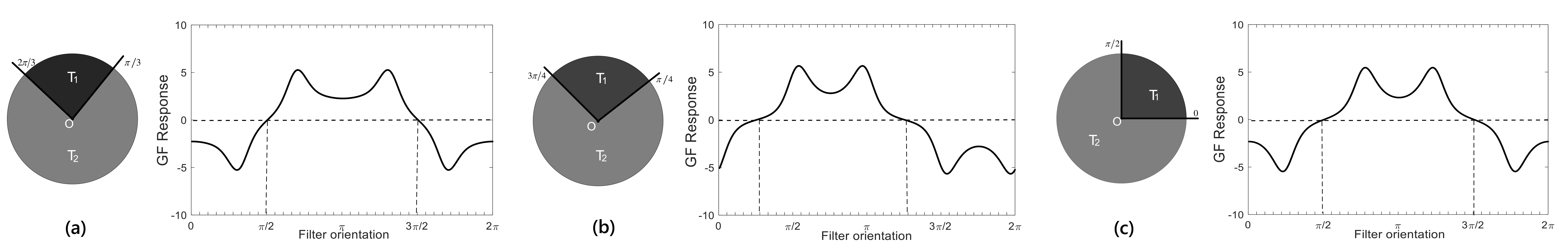}
	\caption{Affine transformation of L-type corner model (gray value $T_{1} = 50, T_{2} = 100$). (a) Rotation. $T_{1}$ is rotated $\frac{\pi }{2}$ counterclockwise. (b) Zoom. (c) Rotation and zoom. Their corresponding Gabor filtered responses($ f = 0.2,\gamma  = 0.6,\eta  = 1.2 $) are shown behind the appropriate L-type corner model respectively.}
	\label{fig_LtypeTransform}
\end{figure*}

The step edge, Y-shaped corner, L-shaped corner, X-shaped corner, T-shaped corner, and star-shaped corner models are shown in Fig.\ref{fig_corner-models}, along with their corresponding Gabor filtered responses. The responses of the anisotropic Gabor functions are different in different corner models, so we can effectively distinguish the step edge and various corners. At the same time, we also challenge the performance of the Gabor function in the face of an affine transformation. Fig.\ref{fig_LtypeTransform} shows the Gabor filtered responses after affine transformation with an L-type corner. Among them, Fig.\ref{fig_LtypeTransform}(a) makes rotation transformation in Fig.\ref{fig_corner-models}(b), Fig.\ref{fig_LtypeTransform}(b) enlarges the rotated image in Fig.\ref{fig_LtypeTransform}(a), and Fig.\ref{fig_LtypeTransform}(c) includes rotation and magnification transformation. The imaginary part of the Gabor function can still present the unique functional responses of the transformed L-type corners. It can still be distinguished from the step edge and other corner models.
According to the research above, anisotropic Gabor filters may show the intensity change disparities between step edges and corners while maintaining a certain level of robustness in the detection performance. Therefore, it can be used to quickly and effectively detect corners.

\section{Gabor-based Corner Detection}
First, a multi-directional structure tensor based on Gabor features is designed and is extensible in this section. Then a novel algorithm is proposed which achieves corner measurement and detection by analyzing images at multi-scales.

\subsection*{3.1 An Extensible Multi-directional Structure Tensor Based on Gabor Features}
Based on previous research\cite{noble1988finding,zhang2020corner30}, the structure tensor in the horizontal and vertical directions has defects in depicting the differences between step edges and corners, leading to missed or wrong corner detection. In addition, corners are structurally unique points on the image, which have their corresponding angles, so the intensity varies a lot in some directions. Therefore, it is inevitable to describe local structure information in multiple directions.
\begin{figure*}[b]
	\hrulefill
	\begin{align*}\label{eq_EtM}
	E\left( {\Delta t} \right) &\approx \frac{1}{{K{{\left( {n + 1} \right)}^2}}}\sum\limits_{v =  - \frac{n}{2}}^{ - \frac{n}{2}} {\sum\limits_{w =  - \frac{n}{2}}^{\frac{n}{2}} {r\left( {v,w} \right)} } {\left( {\left[ {\begin{array}{*{20}{c}}
				{GaborI{}_{f,{\theta _1}}}&{GaborI{}_{f,{\theta _2}}}& \ldots &{GaborI{}_{f,{\theta _K}}}
				\end{array}} \right]{{\left[ {\begin{array}{*{20}{c}}
						{\Delta t}&{\Delta t}& \ldots &{\Delta t}
						\end{array}} \right]}^{\rm T}}} \right)^2}
	\nonumber\\&= \frac{1}{{K{{\left( {n + 1} \right)}^2}}}\left[ {\begin{array}{*{20}{c}}
		{\Delta t}&{\Delta t}& \ldots &{\Delta t}
		\end{array}} \right]M{\left[ {\begin{array}{*{20}{c}}
			{\Delta t}&{\Delta t}& \ldots &{\Delta t}
			\end{array}} \right]^{\rm T}}
	\tag{12}
	\end{align*}	
	\begin{align*}\label{eq_M}
	M = r(v,w)\left[ {\begin{array}{*{20}{c}}
		{\sum\limits_{v =  - \frac{n}{2}}^{\frac{n}{2}} {\sum\limits_{w =  - \frac{n}{2}}^{\frac{n}{2}} {GaborI_{f,{\theta _1}}^2} } }& \cdots &{\sum\limits_{v =  - \frac{n}{2}}^{\frac{n}{2}} {\sum\limits_{w =  - \frac{n}{2}}^{\frac{n}{2}} {Gabor{I_{f,{\theta _1}}}Gabor{I_{f,{\theta _K}}}} } }\\
		\vdots & \ddots & \vdots \\
		{\sum\limits_{v =  - \frac{n}{2}}^{\frac{n}{2}} {\sum\limits_{w =  - \frac{n}{2}}^{\frac{n}{2}} {Gabor{I_{f,{\theta _K}}}Gabor{I_{f,{\theta _1}}}} } }& \ldots &{\sum\limits_{v =  - \frac{n}{2}}^{\frac{n}{2}} {\sum\limits_{w =  - \frac{n}{2}}^{\frac{n}{2}} {GaborI_{f,{\theta _K}}^2} } }
		\end{array}} \right]
	\tag{13}
	\end{align*}
	\hrulefill
\end{figure*}

Within this study, we construct a new multi-directional structure tensor based on Gabor features, which can be applied to corner detection algorithms. The principle is mainly to move the detection patch on the input test image, to obtain the structural information on the image, and then analyze and finely mine the local feature information in the detection patch, which contains the gradient changes in each direction and scale, and finally reach the detection of corners. For example, the detected image is ${I\left( {x,y} \right)}$. When the detection patch is moving, let the center point of the region be located by $\left( {x,y} \right)$, the detection patch will move $\Delta t$ in multiple directions around it, and then calculate the intensity change $E\left( {\Delta t} \right)$. Its response can be defined as
\begin{align}\label{eq_Et}
E\left( {\Delta t} \right)& = \frac{1}{{K{{\left( {n + 1} \right)}^2}}}\sum\limits_{p =  - \frac{n}{2}}^{\frac{n}{2}} {\sum\limits_{q =  - \frac{n}{2}}^{\frac{n}{2}} {\sum\limits_{k = 1}^K{m\left( {p,q} \right)} } } \nonumber\\ &{\left[ {I\left( {x + \Delta t\cos {\theta _k},y + \Delta t\sin {\theta _k}} \right) - I\left( {x,y} \right)} \right]^2}
\end{align}
where a square patch is centered at $\left( {x,y} \right)$, ${\theta _k}$ represents the angle between the location of the $k$-th $\Delta t$ and the polar axis, $r\left( {v,w} \right)$ is a circular mask, and $K$ denotes the directions' number. To facilitate the calculation, the Taylor expansion is used to approximate $I$ in Eq.\ref{eq_Et}, and the approximation formula follows
\begin{align}\label{eq_Taylor}
I\left( {x + \Delta t\cos {\theta _k},y + \Delta t\sin {\theta _k}} \right) \approx I\left( {x,y} \right) + \Delta t{I_k}\left( {x,y} \right)
\end{align}
where ${I_k}\left( {x,y} \right)$ represents the directional derivative of $I\left( {x,y} \right)$ in the direction of ${\theta _k}$. Substituting approximation Eq.\ref{eq_Taylor} into Eq.\ref{eq_Et} yields
\begin{align}\label{eq_EtTaylor}
E\left( {\Delta t} \right) \approx \frac{1}{{K{{(n + 1)}^2}}}\sum\limits_{v =  - \frac{n}{2}}^{\frac{n}{2}} {\sum\limits_{w =  - \frac{n}{2}}^{\frac{n}{2}} {\sum\limits_{k = 1}^K {r\left( {v,w} \right){{\left[ {\Delta t{I_k}\left( {x,y} \right)} \right]}^2}} } }
\end{align}
\begin{align}\label{img}
{I_k}(x,y) \simeq Gabor{I_{f,{\theta _k} }}
\end{align}

Therefore, Eq.\ref{img} is brought into Eq.\ref{eq_EtTaylor} to obtain Eq.\ref{eq_EtM}. The multi-directional structure tensor $M$ at multi-scales is represented by Eq.\ref{eq_M}, where $M$ is the product of a circular mask $r\left( {v,w} \right)$ and a $K \times K$ symmetric matrix.

\subsection*{3.2 Detection and Measurement Algorithm}
In this part, we will discuss a new corner measurement method as well as a new corner detection method.

Within this study, a $K \times K$ multi-directional structure tensor $M$ at multi-scales is used, and $K$ eigenvalues $\left\{ {{\lambda _1},{\lambda _2},\cdots,{\lambda _K}} \right\}$ at each scale constitute a new corner measurement to distinguish the input image corners and other points. The proposed corner measurement is denoted by
\begin{align}\label{eq_measure}
{\varsigma _s} = \frac{{\prod\limits_{k = 1}^K {{\lambda _k}} }}{{\sum\limits_{k = 1}^K {{\lambda _k}}  + \varrho }}
\tag{14}
\end{align}
where $\varrho$ represents a minimal constant term ($\varrho  = 2.22 \times {10^{ - 16}}$) that prevents the denominator from appearing as a structural tensor of zero, the pixel $\left( {x,y} \right)$ of the image is identified as a corner if the matching ${\varsigma _s}$ is a maximum within a $\left( {p + 1} \right) \times (q + 1)$ local region and is greater than a threshold ${T_h}$ at each scale ${f_s}$ ($s = 1,2,3$).

Therefore, this leads to our proposed corner detection algorithm, based on the steps of first smoothing the input test image $I\left( {x,y} \right)$ with the Gabor filter and convolving to calculate the $Gabor{I_{f,{\theta_k}}}$ of each pixel in the image. Then we use the Gabor features to construct a novel multi-directional structure tensor $M$. Finally use the multi-scale corner detection function to extract the corners in each detection patch. In this paper the specific steps of the corner detection algorithm are as follows:
\begin{itemize}
	\item [1)]
	Smooth the test image with the multi-scales Gabor filters that are multi-directionally anisotropic and deduce the
	multi-directional Gabor filtered responses at multi-scales as
	in Eq.\ref{eq_GaborI}.
	\item [2)]
	For each image's pixel $\left( {x,y} \right)$, the matrix $M$ is constructed at multi-scales, as shown in Eq.\ref{eq_M}.
	\item [3)]
	The $K$ eigenvalues $\left\{ {{\lambda _1},{\lambda _2},\cdots,{\lambda _K}} \right\}$ on each scale are calculated as in Eq.\ref{eq_measure}.
	\item [4)]
	A pixel is marked as a candidate corner if its corresponding corner measure is a maximum within a $\left( {p + 1} \right) \times (q + 1)$ local region and is greater than the minimum scale threshold $T_{h}$.
	\item [5)]
	The candidate corner is marked as a corner when its corresponding corner measurement is greater than the $T_{h}$ threshold on each scale.
\end{itemize}

\section{Experimental Results and Performance Evaluation}
In this section, we perform four comprehensive experiments to evaluate the proposed detector from different perspectives applied in other domains. Our proposed corner detection method is compared with the state-of-the-art corner detection methods. First, ground truth images (one complex and one simple scene) commonly used for the experiments are selected to test the detectors. The six detectors is compared by evaluation metrics such as detection accuracy and localization error. Then 30 test images with real-world scenes are selected from the\cite{balntas2017hpatches} to compare the average repeatability of 13 detectors (nine conventional detectors and four deep learning-based detectors) under 6510 different transformed image conditions. In addition, image matching and 3D reconstruction require the important aspect of extracting feature points from images. Thus, our experiments combine HardNet++\cite{mishchuk2017working} and SIFT\cite{lowe2004distinctive} descriptors with various detectors, respectively. In codes the parameter settings for the proposed detector are: $f \in \left\{ {0.15,0.2,0.25} \right\},\ K=6,\ \left( {p + 1} \right) \times \left( {q + 1} \right) = 15 \times 15,$ and ${T_h} = 2 \times {10^8}$.

\subsection*{4.1 Evaluation of Detection Performance Based on Ground Truth Images}
False corners and missed corners occur in the process of image corner extraction by the detectors. We evaluate the performance of the detectors in this experiment. Let the set $P_{GT}=\left \{ \left (x _{t},y_{t} \right ),\ t=1,2,\cdots, N_{1} \right \}$ be the corners of the ground truth image and the set $P_{DC}=\left \{ \left (\hat{x_{t}},\hat{y_{t}} \right),\ t=1,2,\cdots ,N_{2} \right \}$ be the corners extracted from a input original image by a corner detector. If a corner $\left ( x_{t},y_{t} \right )$ in the set $P_{DC}$, corresponding to the set $P_{GT}$, can be found with a value less than the set threshold $\tau$ (here $\tau=4$), then the corner $\left ( x_{t},y_{t} \right )$ is detected correctly. Otherwise, the corner $\left (\hat{x_{t}},\hat{y_{t}}\right)$ in the set $P_{DC}$ is considered a false corner. For a corner $\left ( x_{t},y_{t} \right )$ in the set $P_{GT}$, the corner $\left ( x_{t},y_{t} \right )$ in the set $P_{DC}$ is marked as a missed corner if it cannot be found in the set $P_{DC}$ with less than the set threshold $\tau$, indicating that the corner $\left ( x_{t},y_{t} \right )$ in the set $P_{GT}$ is not detected accurately. Then the corner is marked as a missed corner. In addition, the corner localization error represents the average distance of all successfully matched corner pairs $P_{MC}$ ($P_{MC}=\left \{ \left ( \hat{x_{d}},\hat{y_{d}} \right ),\left ( x_{d},y_{d} \right ):\ d=1,2,\cdots ,N_{m} \right \}$). The average corner localization error $L_{ce}$ is evaluated as
\begin{align}\label{local_error}
L_{ce}=\sqrt{\frac{1}{N_{m}}\sum_{d=1}^{N_{m}}\left ( \left ( \hat{x_{d}}-x_{d} \right )^{2}+\left ( \hat{y_{d}}-y_{d} \right )^{2} \right )}
\tag{15}
\end{align}

\begin{figure}[h]
	\centering	
	\includegraphics[width=3.5in,height=1.3in]{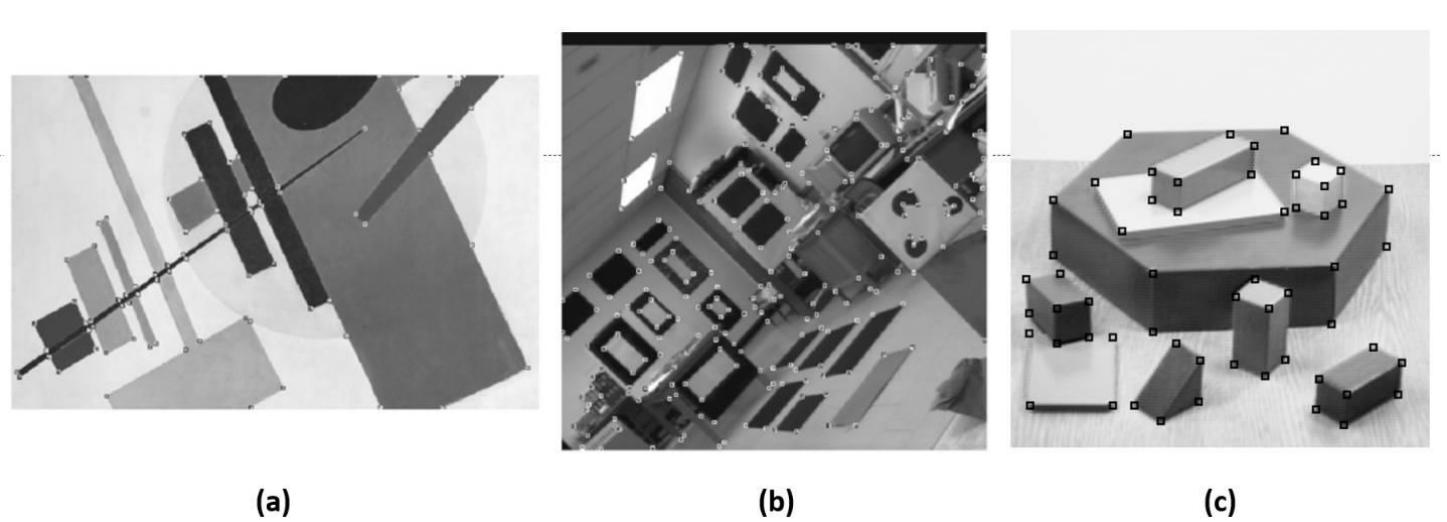}	
	\caption{Test images. (a) `Lab' with GTs, (b) `Geometric' with GTs, (c) 'Block' with GTs.}
	\label{fig_standardGroundtruth}
\end{figure}

\begin{figure*}
	\centering
	\includegraphics[width=6.4in]{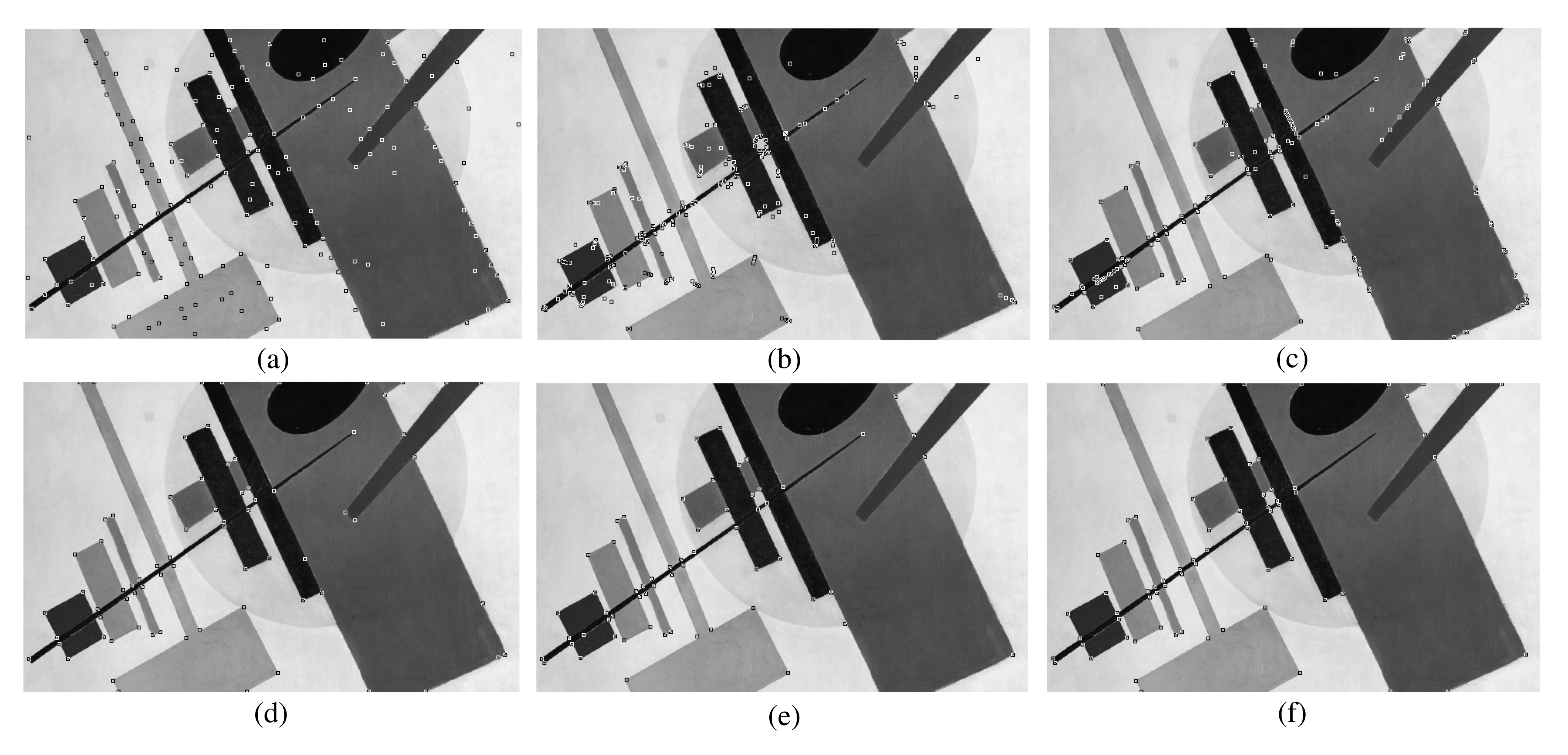}
	\caption{The detection results of the proposed method and the compared corner detectors on the image `Geometric'. (a) Harris\cite{harris1988combined}, (b) Harris-Laplace\cite{mikolajczyk2004scale}, (c) FAST\cite{rosten2008faster}, (d) IPGF\cite{kumar2002defect}, (e) SOGGDD\cite{zhang2020corner30}, (f) proposed method.}
	\label{fig_geometric}
\end{figure*}

\begin{figure*}
	\centering
	\includegraphics[width=6.4in]{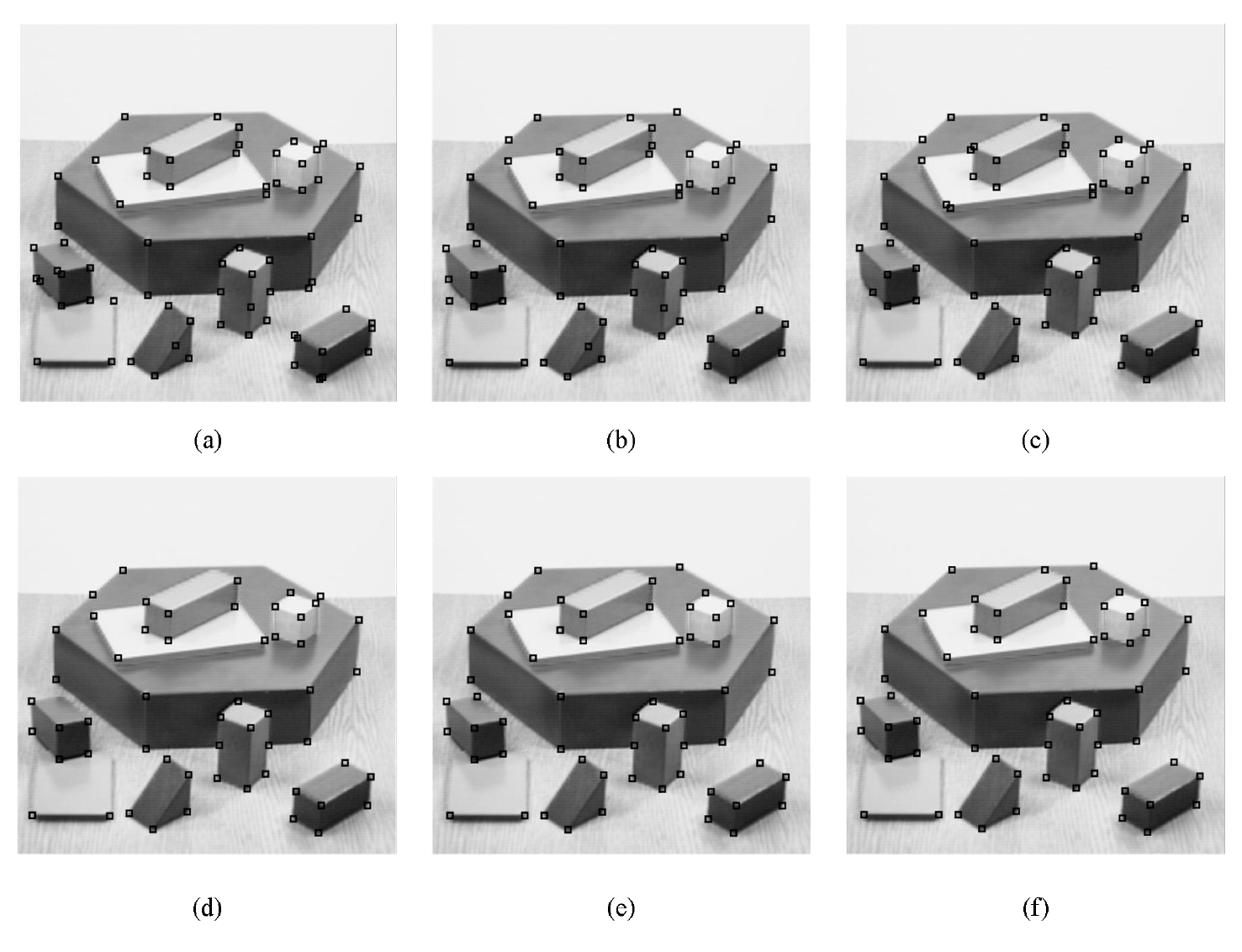}
	\caption{The detection results of the proposed method and the compared corner detectors on the image `Block'. (a) Harris\cite{harris1988combined}, (b) Harris-Laplace\cite{mikolajczyk2004scale}, (c) FAST\cite{rosten2008faster}, (d) IPGF\cite{kumar2002defect}, (e) SOGGDD\cite{zhang2020corner30}, (f) proposed method.}
	\label{fig_kp}
\end{figure*}

\begin{figure*}
	\centering
	\includegraphics[width=6.4in,height=4.5in]{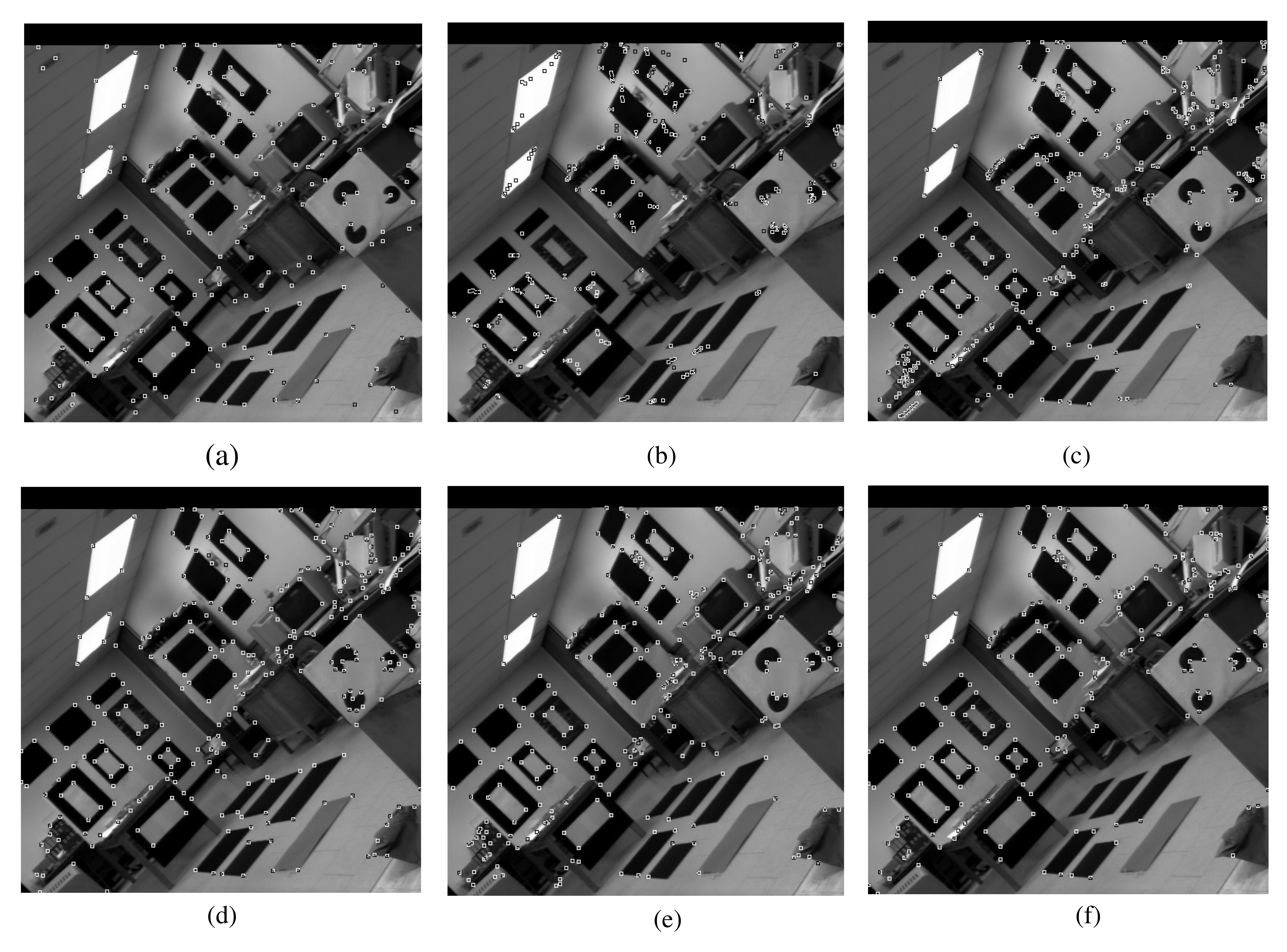}
	\caption{The detection results of the proposed method and the compared corner detectors on the image `Lab'. (a) Harris\cite{harris1988combined}, (b) Harris-Laplace\cite{mikolajczyk2004scale}, (c) FAST\cite{rosten2008faster}, (d) IPGF\cite{kumar2002defect}, (e) SOGGDD\cite{zhang2020corner30}, (f) proposed method.}
	\label{fig_lab}
\end{figure*}

In this experiment, we use three standard test images (i.e., 'Geometry', 'Block' and 'Lab'), shown in Fig.\ref{fig_standardGroundtruth}, for their corresponding ground truths. The ground truths of images 'Geometry', 'Block' and 'Lab' contain 84 corners , 57 corners and 249 corners, respectively. The parameters of each detector are adjusted initially to obtain the best detection results for all detectors on both standard test images. Then our proposed detector and the other five detectors (Harris\cite{harris1988combined}, Harris-Laplace\cite{mikolajczyk2004scale}, FAST\cite{rosten2008faster}, IPGF\cite{kumar2002defect}, and SOGGDD\cite{zhang2020corner30}) are used for corner detection on the three standard test images. The experimental results are shown in Fig.\ref{fig_geometric} , Fig.\ref{fig_kp}, and Fig.\ref{fig_lab}. The missed corners, false corners, and average localization errors are finally experimentally derived and presented in Table.\ref{tab_Groundevalue}.

The detected corners in Fig.\ref{fig_geometric}, Fig.\ref{fig_kp} and Fig.\ref{fig_lab} of the experimental results show that different detectors have different abilities to analyze and extract the corners of the same image. In Fig.\ref{fig_geometric}(a), Fig.\ref{fig_kp}(a), and Fig.\ref{fig_lab}(a), we find that the Harris detector detects the edges of objects as corners because the Harris detector mainly analyzes the structural information in the image's local horizontal and vertical directions. In comparison, the Harris-Laplace detector adds a multi-scale approach to the Harris detector so that the Harris-Laplace detector does not incorrectly detect edges as corners. In Fig.\ref{fig_geometric}(b), Fig.\ref{fig_kp}(b), and Fig.\ref{fig_lab}(b), the Harris-Laplace detector detection results are shown. The false detection rate is reduced because the detector adds Gaussian smoothing. The FAST detector is obtained by modifying the SUSAN detector\cite{smith1997susan}, which increases the size of the circular template to improve the detection accuracy. In Fig.\ref{fig_geometric}(c), Fig.\ref{fig_kp}(c), and Fig.\ref{fig_lab}(c), the FAST detector performs superiorly in flat regions and objects edges, but there are false corners detections in the image's textured areas. In Fig.\ref{fig_geometric}(d), Fig.\ref{fig_kp}(d), and Fig.\ref{fig_lab}(d), the IPGF detector performs better than the previous three detectors when the edge contours in the detected region are clear. For the SOGGDD detector, the corners are judged and extracted by calculating the second-order derivatives of the local regions as presented in Fig.\ref{fig_geometric}(e), Fig.\ref{fig_kp}(e), and Fig.\ref{fig_lab}(e). However, it adds Gaussian filtering when processing the image and loses some smaller areas to the original features, making it difficult to detect accurate corners. As shown in Fig.\ref{fig_geometric}(f), Fig.\ref{fig_kp}(f), and Fig.\ref{fig_lab}(f), our proposed detector, takes the multi-directional structure tensor to obtain the structure information by moving the detection surface slice in the input image and then combines multi-scale and multi-directional anisotropy to inscribe the intensity variation difference between local step edges and corners to detect corners. Compared with other detectors, our detector effectively reduces the number of false corners and misses corners, which improves the detector's capability.

The false and missed corners of the image directly reflect the performance of the detector. Then the total number can be calculated to compare the ability of the detectors to detect the corners of the images. In Table.\ref{tab_Groundevalue}, the total number of incorrect corners (false and missed corners) calculated for the six detectors 'Geometric'  images are 157, 176, 109, 18, 26, and 15, respectively. In addition, Table.\ref{tab_Groundevalue} also shows the average corner localization error calculated by Eq.\ref{local_error}, which is used as an essential basis for evaluating corner detectors. Our detector's average corner localization error obtains the best performance on the three test images.

\begin{table*}
	\begin{center}
		\caption{Performance comparison for the six detectors on three test images with ground truth.}
		\label{tab_Groundevalue}
		\renewcommand\arraystretch{1.3}
		\resizebox{\textwidth}{!}{
		\begin{tabular}{cccclccclccc}
			\hline
			\multirow{2}{*}{Detectors} & \multicolumn{3}{c}{Test image 'Geometric'}                                                               &  & \multicolumn{3}{c}{Test image 'Block'}                                                                   &  & \multicolumn{3}{c}{Test image 'Lab'}                                                                      \\
			\cline{2-4}\cline{6-8}\cline{10-12}
			& Missed corners & False coeners & \begin{tabular}[c]{@{}c@{}}Localization error\\(in pixels)\end{tabular} &  & Missed corners & False coeners & \begin{tabular}[c]{@{}c@{}}Localization error\\(in pixels)\end{tabular} &  & Missed corners & False coeners & \begin{tabular}[c]{@{}c@{}}Localization error\\(in pixels)\end{tabular}  \\
			\hline
			Harris                     & 38             & 119           & 1.216                                                                   &  & 2              & 13            & 1.270                                                                   &  & 110            & 65            & 1.479                                                                    \\
			Harris-Laplace             & 150            & 26            & 1.358                                                                   &  & 2              & 8             & 1.028                                                                   &  & 143            & 232           & 2.508                                                                    \\
			FAST                       & 16             & 93            & 1.200                                                                   &  & 4              & 9             & 1.195                                                                   &  & 75             & 169           & 1.581                                                                    \\
			IPGF                       & 14             & 4             & 1.201                                                                   &  & 6              & 3             & \textbf{0.984}                                                                   &  & 55             & 93            & 1.640                                                                    \\
			SOGGDD                     & 25             & 1             & 1.033                                                                   &  & 3              & 3             & 1.128                                                                   &  & 68             & 105           & 1.957                                                                    \\
			Proposed                   & 14             & 1             & \textbf{1.028}                                                          &  & 2              & 2             & 1.034                                                                   &  & 92             & 72            & 1.556                                                                    \\
			\hline
		\end{tabular}}
	\end{center}
\end{table*}

\subsection*{4.2 Repeatability under Affine Transformation}
In this experiment, we use the metric of average repeatability\cite{awrangjeb2008robust_cpdaevaluation} to evaluate the robustness of the corner detectors. The original and the transformed image are used as input images, and then the corners are detected using the detectors, and the corresponding repeatability is finally calculated.

\begin{figure*}
	\centering
	\includegraphics[width=7.0in]{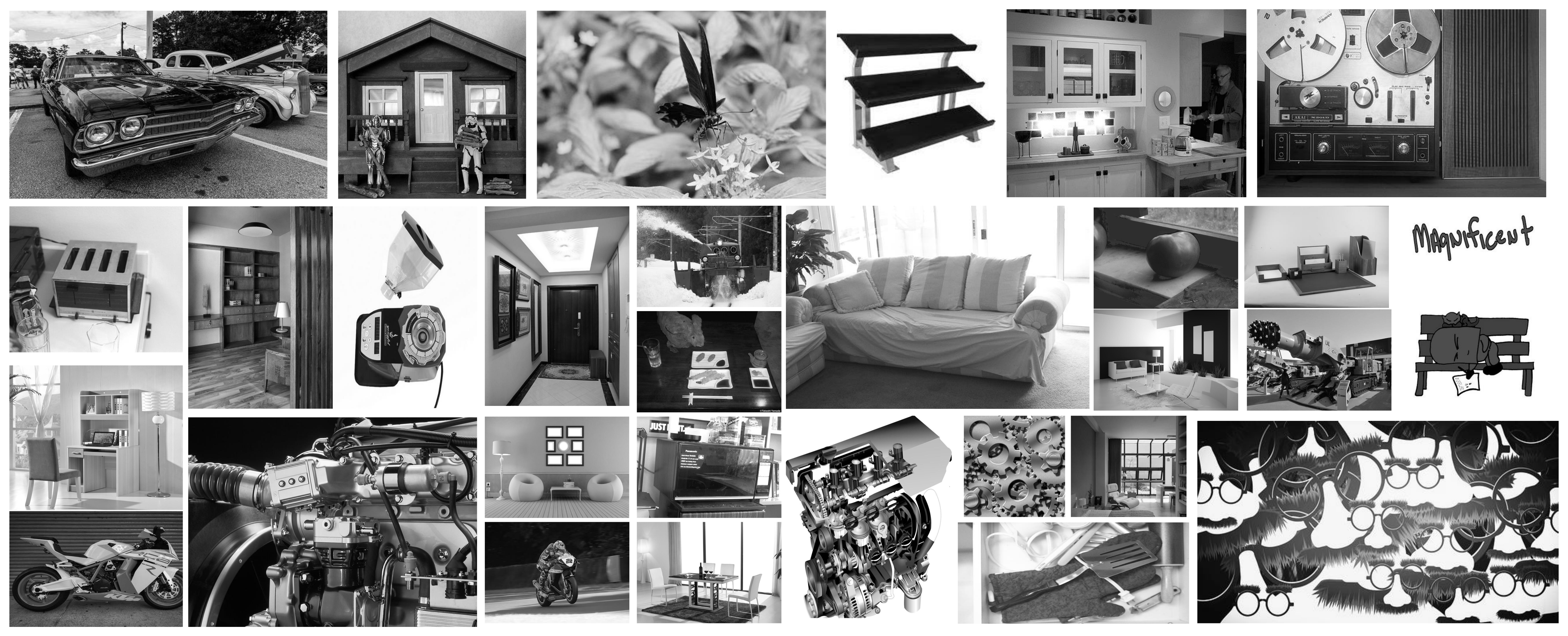}
	\caption{Test images.}
	\label{fig_evaluation_Pics}
\end{figure*}

The average repeatability of the detectors is measured using thirty original images with varied scenes, as shown in Fig.\ref{fig_evaluation_Pics}. We have a total of 6,510 transformed test images, which are obtained by applying each original image to one of the six transformations listed below:
\begin{itemize}
	\item Rotation: The original images are rotated within the range of $\left [ -\frac{\pi }{2},\frac{\pi }{2} \right ]$ with $10^{\circ}$ apart to produce rotated images, excluding $0^{\circ}$.
	\item Uniform scaling: The uniform scaled images are created by scaling the original images between $\left [0.5,2 \right ]$ at 0.1 intervals, excluding 1.
	\item Non-uniform scaling: Scaling the original images in the horizontal and vertical directions with 0.1 spacing within the ranges $\left [ 0.7,1.5 \right ]$ and $\left [ 0.5,1.8 \right ]$ produces non-uniform scaled images.
	\item Shear transformations: The original images are sheared in the range of $p=\left [ -1,1 \right ]$ at an interval of 0.1 to obtain the sheared images, excluding 0, and the transformation formula is shown below
	\begin{align}
	\left[ {\begin{array}{*{20}{c}}
		{w'}\\
		{h'}
		\end{array}} \right] = \left[ {\begin{array}{*{20}{c}}
		1&p\\
		0&1
		\end{array}} \right]\left[ {\begin{array}{*{20}{c}}
		w\\
		h
		\end{array}} \right].\nonumber
	\end{align}
	\item Lossy JPEG compression: The original images are compressed to acquire the compressed images with compressed quality, and the compression factor is in $\left [ 5,100 \right ]$ with five apart.
	\item Gaussian noise: Add zero-mean Gaussian white noise to the original image and the standard deviation value of the noise $\left [ 1,15 \right ]$, interval 1.	
\end{itemize}

Awrangjeb proposed the average repeatability in\cite{awrangjeb2008robust_cpdaevaluation}, which is used to measure the number of repetitions of corners between the original image and the affine transformed image as a means of comparing the performance of the detectors. The average repeatability $\omega_{avg}$ is calculated as
\begin{align}
\omega_{avg}=\frac{D_{r}}{2}\left ( \frac{1}{D_{ip}}+\frac{1}{D_{it}} \right )
\tag{15}
\end{align}
where $D_{ip}$ denotes the corners' number extracted from the primary image, $D_{it}$ means the corners' number extracted from the transformed image, and $D_{r}$ represents the number of corners matched by $D_{ip}$ and $D_{it}$. When a corner $(x_{t},y_{t})$ is detected by the detector in the transformed image, which is close to a corner $(x_{p},y_{p})$ location from the original image, a feature matching pair is created (within $2\times2$ pixels), then a repeated corner is obtained. The detector's performance is better if the average repetition rate is higher.

\begin{figure}
	\centering
	\includegraphics[width=3.7in]{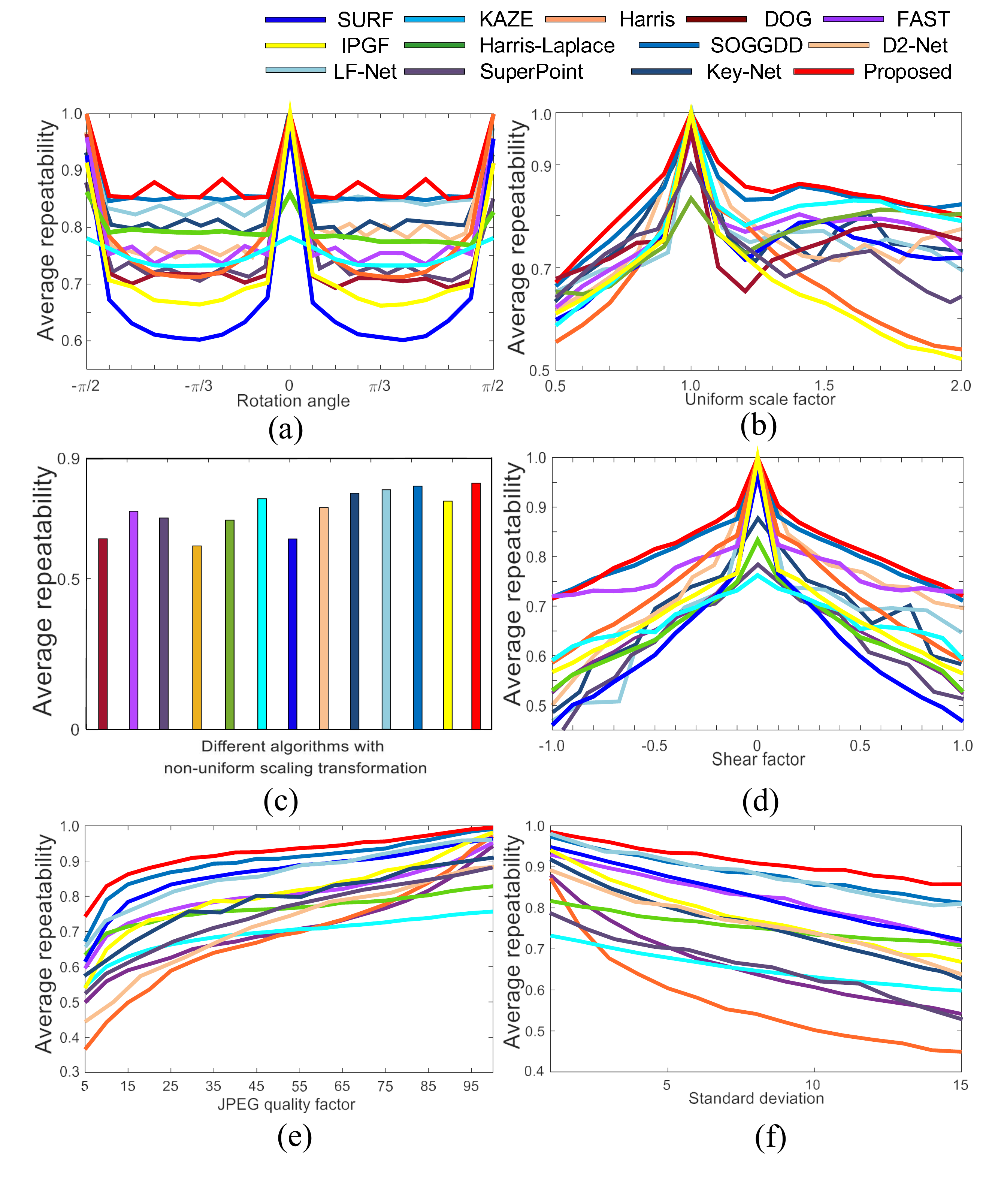}
	\caption{Average repeatabilities of the thirteen detectors under rotation, uniform scaling, non-uniform scaling, shear transforms, lossy JPEG compression, and zero-mean white Gaussian noises.}
	\label{fig_evalu_CPDA}
\end{figure}

The parameters of the thirteen detectors were set to default values by calculating the evaluation metric of repetition rate as described above. As shown in Fig.\ref{fig_evalu_CPDA}, our proposed detector was compared with twelve detectors, and the highest average repeatability was obtained in various transformations of the test image detection.

\subsection*{4.3 Image Matching and Matching Score under the VLBenchmarks}
We combine the six corner detectors with the HardNet++ descriptor\cite{mishchuk2017working} to establish the matching of two image feature points in this experiment. The data used for the experiment is the HPatch dataset\cite{balntas2017hpatches} (consisting of 59 scenes for viewpoint alterations and 57 scenes for illumination alterations), which is detected by 18 detectors used to evaluate the matching scores under the VLBenchmarks\cite{lenc12vlbenchmarks}.

VLBenchmarks is a scoring mechanism for image matching presented in\cite{mikolajczyk2005comparison}. Calculating the matching score consists of two steps. Firstly, the overlap error of the matched two areas extracted from the primary and altered images is calculated. The two areas are judged as a area if the overlap error $\rho$ is sufficiently less than the set threshold value. For repeatability evaluation\cite{mikolajczyk2005comparison} of regions in two images, the overlap error $\rho$ is
\begin{equation}
\rho=1 - \frac{A\cap{Q^{T}}BQ}{A\cup{Q^{T}}BQ}
\tag{16}
\end{equation}
where $A$ denotes a area of the primary image, $B$ denotes a corresponding area of the altered image, $Q$ denotes the corresponding homography matrix of two areas from the primary and the altered images, and $Q^{T}$ is the transpose matrix of $Q$. The area of the altered image $B$ is mapped to the region of the primary image $A$ by the corresponding homography matrix $Q$ to form $A\cup{Q^{T}}BQ$. The value of 1 minus the ratio of the union set $A$, and the intersection set $B$ is the overlap error. When the overlap error value $\rho$ of the two calculated areas is less than the set threshold ($\iota=0.4$), then the two areas are considered similar, and correspondence is established.

Secondly, the matching score is calculated as
\begin{equation}
PS_n=\frac{PG_{1n}}{min(PG_1,PG_n)}(n=2,...,6)
\tag{17}
\end{equation}
where $PG_{1n}$ denotes the total number of actual matching keypoint pairs (matching points between the primary image and the n-th altered image), $PG_1$ denotes the number of keypoints which is detected in the primary image, and $PG_n$ denotes the number of keypoints that are detected in the $n$-th altered image.

\begin{figure}
	\centering
	\includegraphics[width=3.6in,height=2.0in]{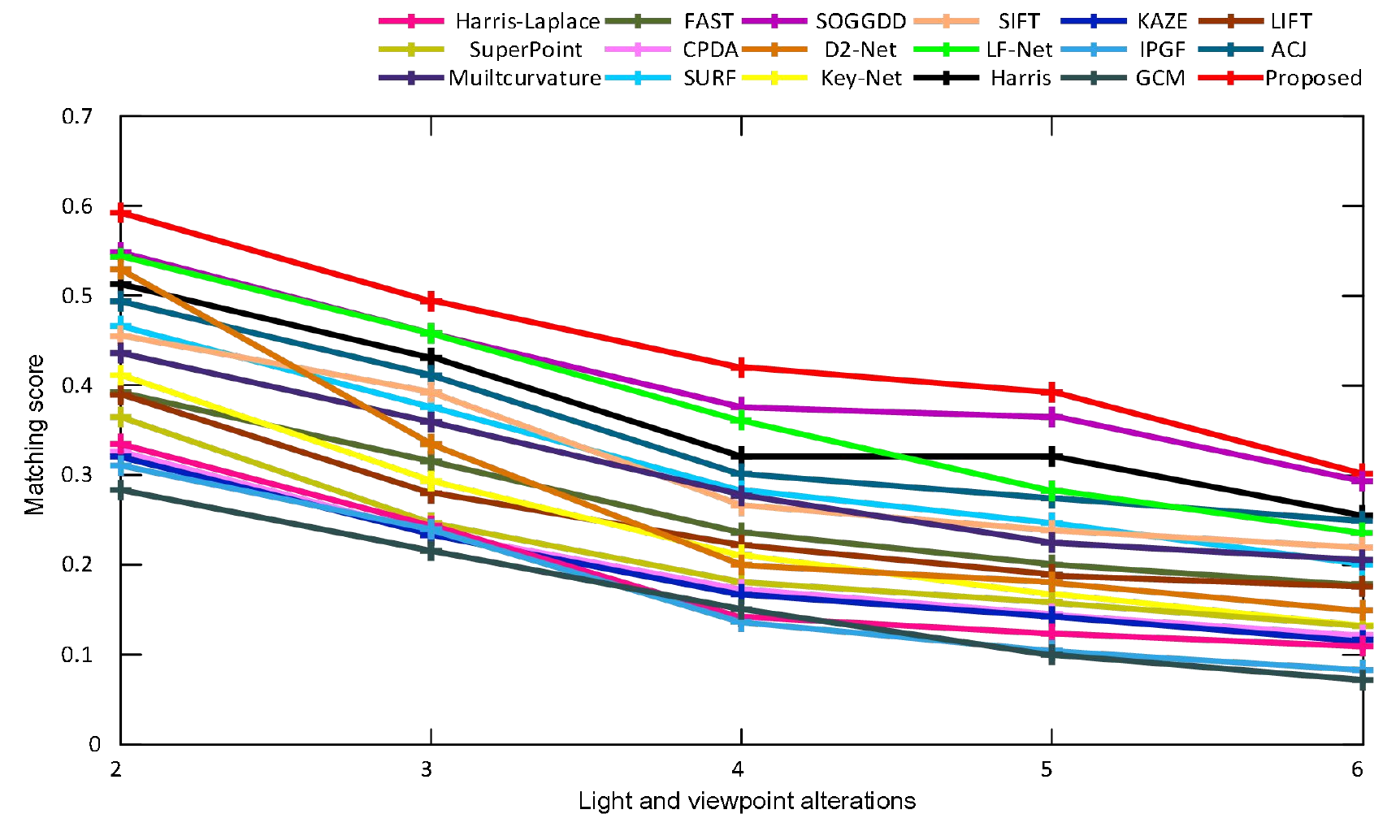}
	\caption{Average matching scores of the eighteen detection methods with the HardNet++ descriptor\cite{mishchuk2017working}.}
	\label{fig_matching_score}
\end{figure}

Therefore, we established the correspondence between two images with differential variations in the same scene and calculated their matching scores. In Fig.\ref{fig_matching_score} the evaluated result of matching scores is demonstrated by 18 detectors (Harris-Laplace\cite{mikolajczyk2004scale}, FAST\cite{rosten2008faster}, SOGGDD\cite{zhang2020corner30}, SIFT\cite{lowe2004distinctive}, KAZE\cite{alcantarilla2012kaze}, LIFT\cite{yi2016lift}, CPDA\cite{awrangjeb2008robust_cpdaevaluation}, SURF\cite{bay2008speeded}, IPGF\cite{kumar2002defect}, Harris\cite{harris1988combined}, ACJ\cite{xia2014accurate29}, GCM\cite{teng2015effective}, Muiltcurvature\cite{zhang2019discrete24}, LF-Net\cite{LF-Net2018}, SuperPoint\cite{2021Unsupervised}, Key-Net\cite{Key-Net2019}, D2-Net\cite{D2-Net2019}, and Proposed). In Fig.\ref{fig_matching_score}, our proposed method performs well or even better than all baselines under this evaluation criteria. Furthermore, as shown in Table.\ref{tab_score}, the average match scores of each detector are quantitatively represented, from which our proposed method improves the scores by 3.04$\%$-23.61$\%$ under light and viewpoint alterations.
\begin{table}
	\centering
	\caption{The average score under light and viewpoint alterations.}
	\label{tab_score}
	\begin{tabular}{ c c }
		\hline
		Detectors & Average matching score \\
		\hline
		Harris-Laplace & 0.2159 \\ FAST & 0.2923 \\ SOGGDD & 0.4216 \\
		SIFT & 0.3180 \\ KAZE & 0.2412 \\ LIFT & 0.2663 \\
		CPDA & 0.2417 \\ SURF & 0.3293 \\ IPGF & 0.2485 \\
		Harris & 0.3864 \\ ACJ & 0.3695 \\ GCM & 0.2638 \\
		Muiltcurvature & 0.3246 \\ LF-Net & 0.3886 \\ SuperPoint & 0.2261 \\
		Key-Net & 0.2642 \\ D2-Net & 0.3071 \\ Proposed & \textbf{0.4520} \\
		\hline
	\end{tabular}
\end{table}

\begin{figure*}
	\centering
	\includegraphics[width=6.4in]{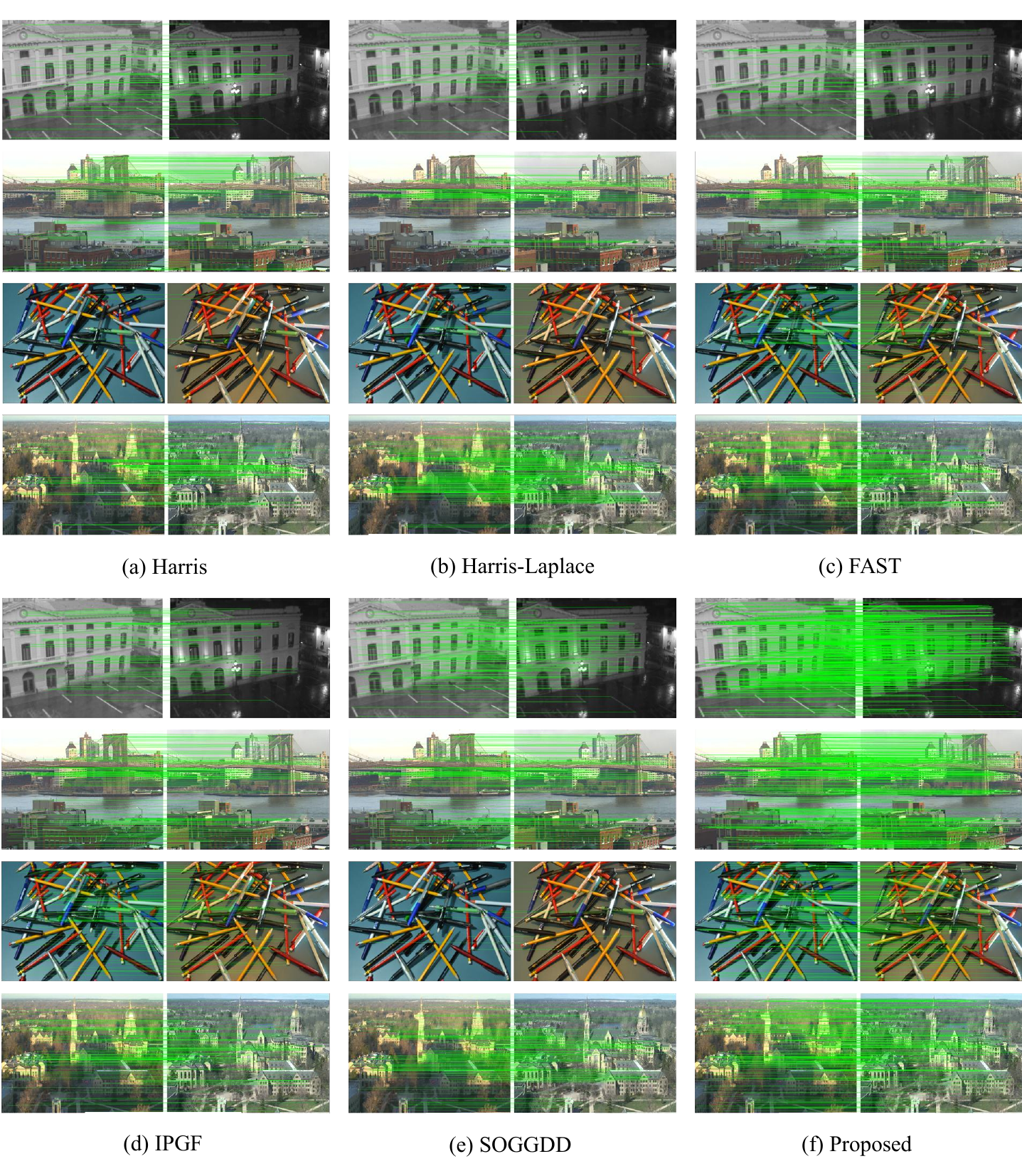}
	\caption{Six detectors combined with HardNet++ descriptors in different lighting are matched with examples of results on the HPatches dataset. The green lines are the pairs of feature points successfully matched.}
	\label{fig_matching_image}
\end{figure*}

\begin{figure*}
	\centering
	\includegraphics[width=6.4in]{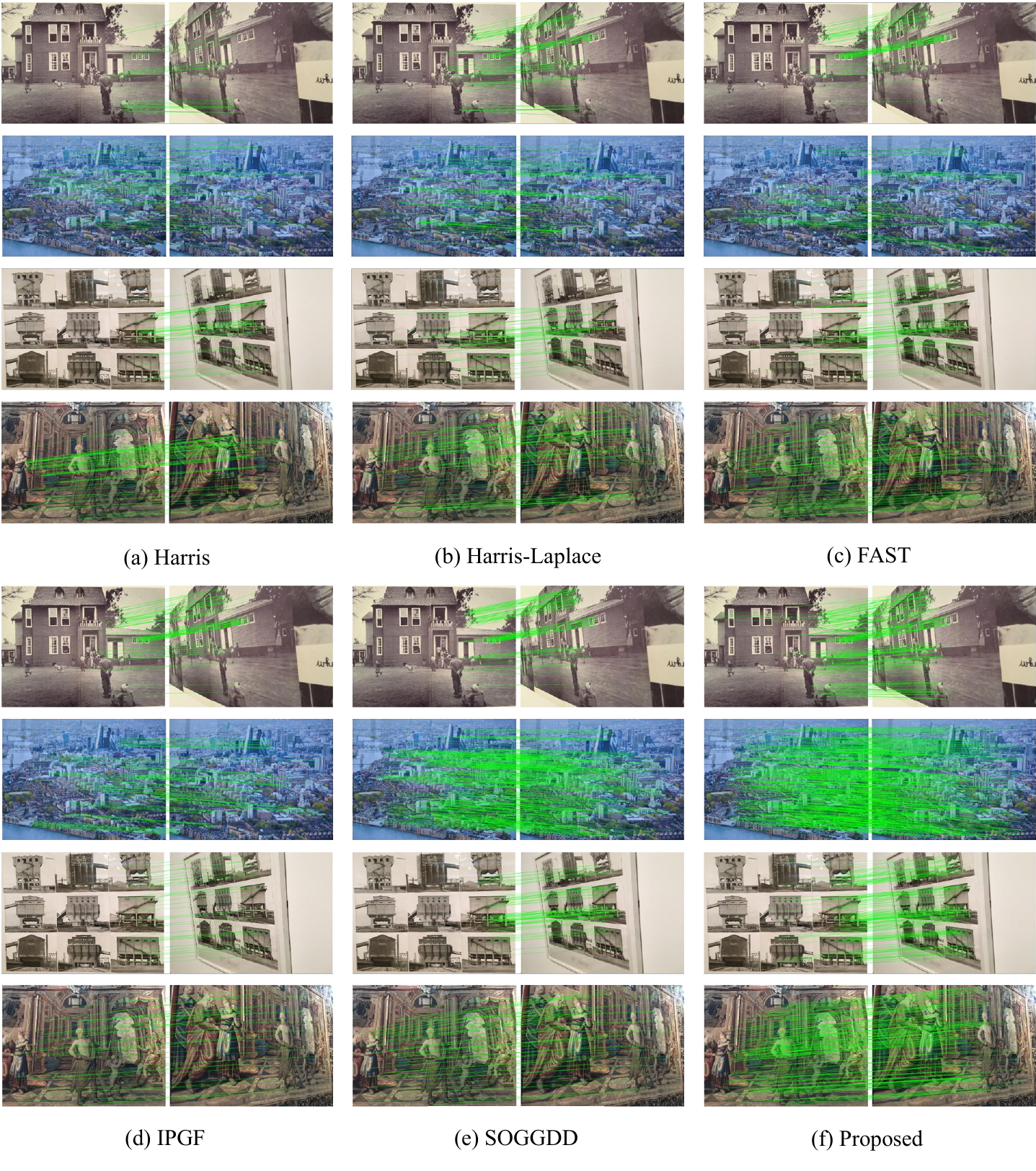}
	\caption{Six detectors combined with HardNet++ descriptors in different viewpoints are matched with examples of results on the HPatches dataset. The green lines are the pairs of feature points successfully matched.}
	\label{fig_matching_image1}
\end{figure*}

Furthermore, five representative detectors (Harris\cite{harris1988combined}, Harris-Laplace\cite{mikolajczyk2004scale}, FAST\cite{rosten2008faster}, IPGF\cite{kumar2002defect}, and SOGGDD\cite{zhang2020corner30}) and our proposed detector with HartNet++ descriptor\cite{mishchuk2017working} are compared. The result of matching is shown in Fig.\ref{fig_matching_image} under the detectors for the four pairs of images, which are the primary images ('Ajuntament', 'Brooklyn', 'Pens', and 'Indiana') and illumination altered images. Fig.\ref{fig_matching_image1} shows the matching result under the detector for the four pairs of images, which are the primary images ('Home', 'London', 'Machines', and 'Yard') and the viewpoint altered images. Our proposed detector shows better dense and accurate matching results than the other five detectors.

\subsection*{4.4 Application to 3D Reconstruction }
We used the proposed detector for image feature point matching in the previous section and obtained significant results. In this experiment, we will now briefly describe their use for 3D reconstruction. More details on applications of these features to reconstruction are available in other papers\cite{moulon2012adaptive3D01}.

The Structure from Motion (SfM) algorithm takes different viewpoint images of an object or scene as input and deduces parameters of the camera by matching between 2D images to reconstruct the point cloud of the object or scene mapping to a three-dimensional coordinate system. In this algorithm, the first step is to detect the feature points of the input image (e.g., SIFT\cite{lowe2004distinctive}) and establish a matching relationship between image pairs, where a miss-match can cause a large error. Robust feature detection and strong feature description will bring significant gains to the final effect, and the general feature correspondence is described in algorithm\ref{algo_matches}. The two-view is used as the basis for reconstruction, and the position of the new view in 3D space is calculated using parameters such as camera position and pose to complete the 3D point cloud.

\begin{algorithm}[h]
	\caption{\small Computation of geometry-consistent pairwise correspondences.}
	\label{algo_matches}
	\small \textbf{Require:} image set.\\
	\small \textbf{Ensure:} pairwise point correspondences that are consistent geometrically. \\
	\small \textbf{Compute putative matches:}
	\begin{enumerate}
		\small \item detect features in each image and build their descriptor;\\
		\small \item match descriptors (using brute force or approximate nearest neighbor);\\
	\end{enumerate}
	\small \textbf{Filter geometric-consistent matches:}
	\begin{enumerate}
		\small \item estimate fundamental matrix F;\\
		\small \item estimate homography matrix H;\\
	\end{enumerate}
\end{algorithm}

\begin{figure*}
	\centering
	\includegraphics[width=6.4in]{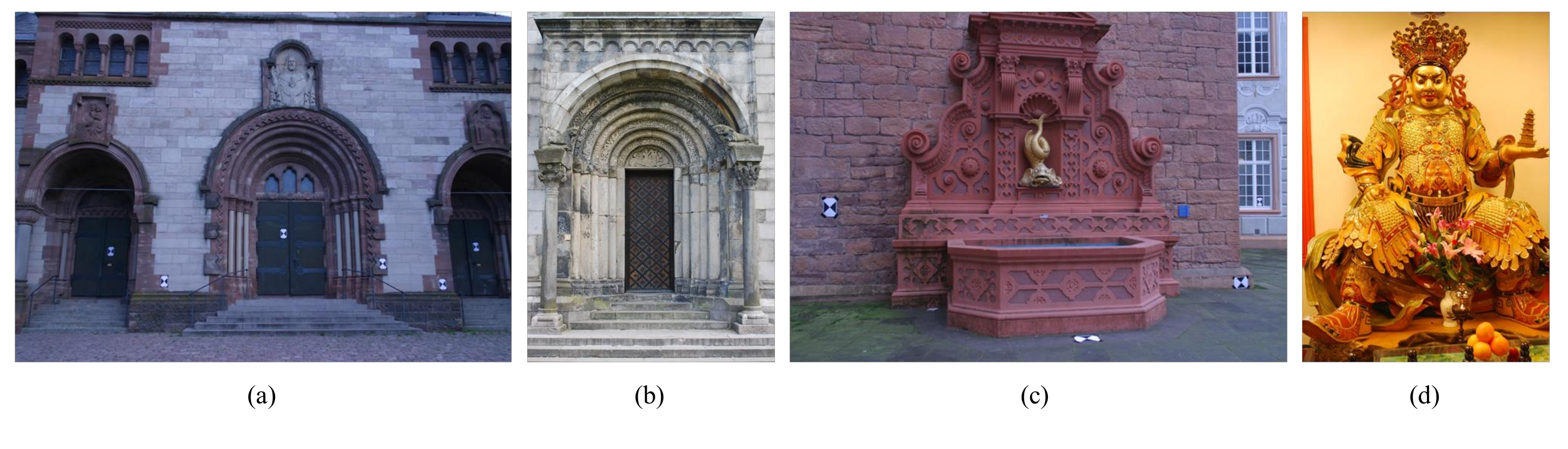}
	\caption{Typical buildings of the four datasets used to perform the 3D reconstruction. (a) Herz-Jesu-P25\cite{strecha2008benchmarking}. (b) Door, Lund\cite{olsson2011stable}. (c) fountain-P11\cite{strecha2008benchmarking}. (d) Golden statue somewhere in Hong Kong\cite{olsson2011stable}. }.
	\label{fig_sfmpictures}
\end{figure*}

\begin{figure*}
	\centering
	\includegraphics[width=6.4in]{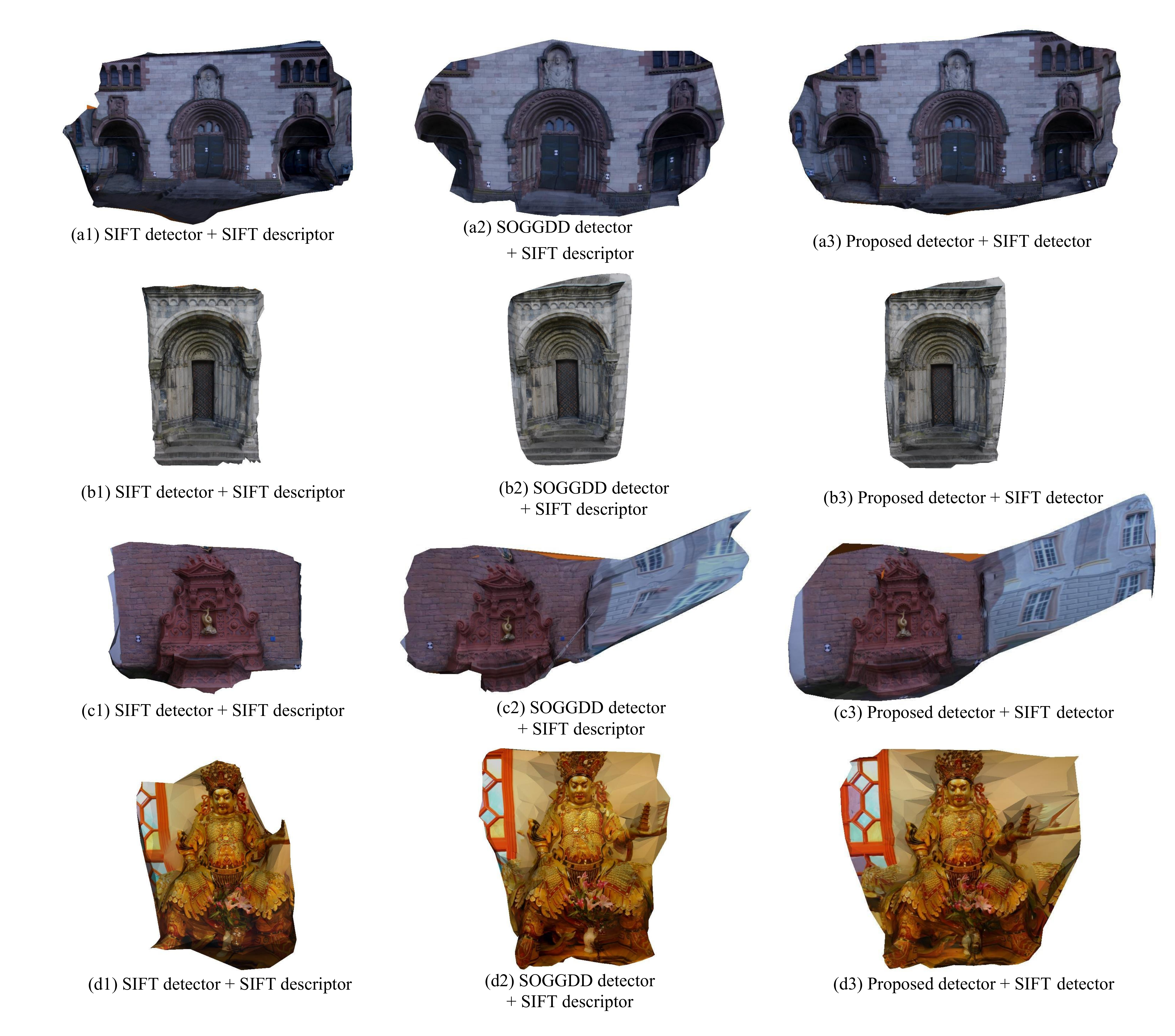}
	\caption{Reconstruction results for four datasets. Every detector catches 3800 feature points per image, and feature struct information is represented using a compatible SIFT descriptor. }
	\label{fig_reconstruction}
\end{figure*}

Due to the robust and accurate characteristics of the proposed detector, the detector combined with the optimized SIFT descriptor was used for 3D reconstruction. Three excellent visual Libraries-OpenCV, OpenMVG, and OpenMVS were used for the operation structure component. Furthermore, four datasets\cite{strecha2008benchmarking,olsson2011stable} shown in Fig.\ref{fig_sfmpictures} are used to perform the 3D reconstruction. In addition, other excellent algorithms(e.g.,  SIFT\cite{lowe2004distinctive}, SOGGDD\cite{zhang2020corner30}) were compared with us in this hierarchy. The comparison results are shown in Fig.\ref{fig_reconstruction}.

The datasets that we use contain rich scenes, Fig.\ref{fig_sfmpictures}(a) and Fig.\ref{fig_sfmpictures}(c) contain 25 and 11 scenes, respectively. These images are frequently used in applications \cite{strecha2008benchmarking} concerning reconstructing a highly particular object or scene. In the first and third rows of Fig.\ref{fig_reconstruction}, we can see the comparison of the reconstruction effects of these two data sets. SOGGDD\cite{zhang2020corner30} and our proposed detector application performs considerably better in reconstruction integrity, but in Fig.\ref{fig_reconstruction}(c2) and Fig.\ref{fig_reconstruction}(c3), our proposed method showed better results. Fig.\ref{fig_sfmpictures}(b) and Fig.\ref{fig_sfmpictures}(d) contain high-resolution images captured from 12 and 18 fixed viewpoints, respectively. The performance of the three detectors is slightly different in the third row of Fig.\ref{fig_reconstruction}. However, in Fig.\ref{fig_sfmpictures}(d), our detector excels in the completeness of the reconstruction and the correct representation of texture features. By observing the 3D reconstruction results, it can be found that our method is used in the reconstruction algorithm, and the reconstruction results contain more information about the scene structure due to the extraction of more accurate corners.

\section{Conclusion}
We evaluate this paper's properties of the Gabor wavelet suppression affine image transform, and the capability of the Gabor filters to discriminate the intensity changes of step edges and several general corners is investigated and obtained, thus proving that the existing Gabor wavelet-based corner detectors using local structure information cannot effectively detect corners. Then, we propose a new corner metric method and a Gabor filter-based corner detection algorithm to address the shortcomings of traditional Gabor analysis in multi-directional transformations, to accurately present the structural information of multiple directions in images. This also provides a multi-directional and multi-scale analysis tool for the Gabor detector as image feature point extraction.

In addition, the high symmetry of the multi-directional structure tensor can analyze the local orientation information and accurately measure the position of corners at the scale, which effectively improves the detection accuracy and reduces the detection error. Finally, we compare with advanced corner detectors in image matching and 3D reconstruction. Our proposed detector can obtain gradient changes of the structure in each direction of the image under viewpoint and illumination transformations, accurately extract local structural features, and thus obtain a high degree of matching, showing the basic feature information in the actual scene in terms of image matching and 3D reconstruction. However, the improved Gabor filter based corner detection method proposed in this paper is computationally demanding, which is often more computationally complex than many multi-scale based detection methods due to the multi-directional structural information of the analyzed image. Therefore, future work is to explore the basic corner types in images with structural anisotropy, improve the computational speed of the detector, and discover a broader prospect for analyzing local structure meters with higher accuracy in feature extraction and matching. Our proposed detector has excellent potential for generalization in computer vision applications.

% if have a single appendix:
%\appendix[Proof of the Zonklar Equations]
% or
%\appendix  % for no appendix heading
% do not use \section anymore after \appendix, only \section*
% is possibly needed

% use appendices with more than one appendix
% then use \section to start each appendix
% you must declare a \section before using any
% \subsection or using \label (\appendices by itself
% starts a section numbered zero.)
%

%\appendices
%\section{Proof of the First Zonklar Equation}
%Appendix one text goes here.

% you can choose not to have a title for an appendix
% if you want by leaving the argument blank
%\section{}
%Appendix two text goes here.

% use section* for acknowledgment
\section*{Acknowledgment}

This work was supported in part by Innovation Capability Support Program of Shaanxi (No.2021TD-29), in part by the Youth Innovation Team of Shaanxi Universities, in part by the National Natural Science Foundation of China (No.62176204) and in part by the Key Research and Development Plan of Shaanxi Province (No.2022GY-066).

% Can use something like this to put references on a page
% by themselves when using endfloat and the captionsoff option.
\ifCLASSOPTIONcaptionsoff
  \newpage
\fi

% trigger a \newpage just before the given reference
% number - used to balance the columns on the last page
% adjust value as needed - may need to be readjusted if
% the document is modified later
%\IEEEtriggeratref{8}
% The "triggered" command can be changed if desired:
%\IEEEtriggercmd{\enlargethispage{-5in}}

% references section

% can use a bibliography generated by BibTeX as a .bbl file
% BibTeX documentation can be easily obtained at:
% http://mirror.ctan.org/biblio/bibtex/contrib/doc/
% The IEEEtran BibTeX style support page is at:
% http://www.michaelshell.org/tex/ieeetran/bibtex/
\bibliographystyle{IEEEtran}
% argument is your BibTeX string definitions and bibliography database(s)
\bibliography{reference.bib}
%
% <OR> manually copy in the resultant .bbl file
% set second argument of \begin to the number of references
% (used to reserve space for the reference number labels box)

%\begin{thebibliography}{1}
%
%\bibitem{IEEEhowto:kopka}
%H.~Kopka and P.~W. Daly, \emph{A Guide to \LaTeX}, 3rd~ed.\hskip 1em plus
%  0.5em minus 0.4em\relax Harlow, England: Addison-Wesley, 1999.
%
%\end{thebibliography}

% biography section
%
% If you have an EPS/PDF photo (graphicx package needed) extra braces are
% needed around the contents of the optional argument to biography to prevent
% the LaTeX parser from getting confused when it sees the complicated
% \includegraphics command within an optional argument. (You could create
% your own custom macro containing the \includegraphics command to make things
% simpler here.)
%\begin{IEEEbiography}[{\includegraphics[width=1in,height=1.25in,clip,keepaspectratio]{mshell}}]{Michael Shell}
% or if you just want to reserve a space for a photo:

% insert where needed to balance the two columns on the last page with
% biographies
%\newpage

%\begin{IEEEbiographynophoto}{Jane Doe}
%Biography text here.
%\end{IEEEbiographynophoto}

% You can push biographies down or up by placing
% a \vfill before or after them. The appropriate
% use of \vfill depends on what kind of text is
% on the last page and whether or not the columns
% are being equalized.

%\vfill

% Can be used to pull up biographies so that the bottom of the last one
% is flush with the other column.
%\enlargethispage{-5in}

% that's all folks
\end{document}